\documentclass[twoside,11pt]{article}

\usepackage{dnd}
\usepackage{times}

\usepackage{url}

\usepackage{morefloats}
\usepackage{comment}
\usepackage{environ}
\usepackage{paralist}
\usepackage{qtree}
\usepackage{graphicx}
\usepackage{latexsym}
\usepackage{multirow}
\usepackage{tabularx}
\usepackage{caption}
\usepackage{subcaption}
\usepackage{bm}
\usepackage{xspace}
\usepackage{paralist}
\usepackage{comment}
\usepackage{enumitem}
\usepackage{url}

\usepackage{verbatim}

\usepackage{array}

\newcommand{\ocrcomment}[1]{}

\newcommand{\hide}[1]{}

\newcommand{\reltype}[1]{{\sc #1}}

\newcommand{\featSet}[1]{{\textsc{#1}}}
\newcommand{\feat}[1]{\textit{#1}}

\newcommand{\featInitiator}[0]{\feat{Initiator}\xspace}
\newcommand{\featFirstMsgPos}[0]{\feat{FirstMsgPos}\xspace}
\newcommand{\featLastMsgPos}[0]{\feat{LastMsgPos}\xspace}
\newcommand{\featMsgCount}[0]{\feat{MsgCount}\xspace}
\newcommand{\featMsgRatio}[0]{\feat{MsgRatio}\xspace}
\newcommand{\featTokenCount}[0]{\feat{TokenCount}\xspace}
\newcommand{\featTokenRatio}[0]{\feat{TokenRatio}\xspace}
\newcommand{\featTokenPerMsg}[0]{\feat{TokenPerMsg}\xspace}
\newcommand{\featAvgRecipients}[0]{\feat{AvgRecipients}\xspace}
\newcommand{\featAvgToRecipients}[0]{\feat{AvgToRecipients}\xspace}

\newcommand{\featInToListPercent}[0]{\feat{InToList\%}\xspace}
\newcommand{\featReplyRate}[0]{\feat{ReplyRate}\xspace}

\newcommand{\featAddPerson}[0]{\feat{AddPerson}\xspace}
\newcommand{\featRemovePerson}[0]{\feat{RemovePerson}\xspace}

\newcommand{\featDanglingReqPercent}[0]{\feat{DanglingReq\%}\xspace}

\newcommand{\featODPCount}[0]{\feat{ODPCount}\xspace}
\newcommand{\featReqActionCount}[0]{\feat{ReqActionCount}\xspace}
\newcommand{\featReqInformCount}[0]{\feat{ReqInformCount}\xspace}
\newcommand{\featInformCount}[0]{\feat{InformCount}\xspace}
\newcommand{\featConventionalCount}[0]{\feat{ConventionalCount}\xspace}

\newcommand{\genderShort}[0]{\featSet{GEN}\xspace}
\newcommand{\genderEnvShort}[0]{\featSet{ENV}\xspace}
\newcommand{\gender}[0]{\featSet{Gender}\xspace}
\newcommand{\genderEnv}[0]{\featSet{GenderEnv}\xspace}

\newcommand{\featSetNgrams}[0]{\featSet{Lexical}\xspace}

\newcommand{\featSetPositional}[0]{\featSet{Positional}\xspace}
\newcommand{\featSetVerbosity}[0]{\featSet{Verbosity}\xspace}
\newcommand{\featSetMetaData}[0]{\featSet{Thread Structure}\xspace}

\newcommand{\featSetDialogActs}[0]{\featSet{Dialog Acts}\xspace}

\newcommand{\featSetOvertDisplayOfPower}[0]{\featSet{Overt Display of Power}\xspace}

\newcommand{\featSetNgramsShort}[0]{\featSet{LEX}\xspace}

\newcommand{\featSetPositionalShort}[0]{\featSet{PST}\xspace}
\newcommand{\featSetVerbosityShort}[0]{\featSet{VRB}\xspace}
\newcommand{\featSetMetaDataShort}[0]{\featSet{THR}\xspace}

\newcommand{\featSetDialogActsShort}[0]{\featSet{DA}\xspace}

\newcommand{\featSetOvertDisplayOfPowerShort}[0]{\featSet{ODP}\xspace}

\newcolumntype{M}{>{\centering\let\newline\\\arraybackslash\hspace{0pt}}X}
\newcolumntype{Y}{>{\centering\arraybackslash}X}

\usepackage{tikz}

\newcommand{\EnronAll}{\textsc{Enron-All}\xspace}

\newcommand{\EnronGIEC}{\textsc{Enron-APGI}\xspace}

\newcommand{\DAReqAction}[0]{\reltype{Request-Action}\xspace}
\newcommand{\DAReqInform}[0]{\reltype{Request-Information}\xspace}

\newcommand{\nocontentsline}[3]{}
\newcommand{\tocless}[2]{\bgroup\let\addcontentsline=\nocontentsline#1{#2}\egroup}

\usepackage{braket}

\newcommand{\skipfornow}[1]{
}

\renewcommand\cite{\citep}
\newcommand{\newcite}{\citet}

\usepackage{booktabs}
\usepackage{amsmath}

\usepackage{lipsum}
\usepackage[english]{babel}
\usepackage{blindtext}

\usepackage{tabularx}
\usepackage{bm}
\usepackage{paralist}

\usepackage[section]{placeins}

\renewcommand\cite{\citep}

\usepackage{subcaption}

\dndheading{8(2)}{2017}{21--55}{Vinodkumar Prabhakaran, Owen Rambow}{10.5087/dad.2017.202}

\ShortHeadings{Dialog Structure Through the Lens of Gender, Gender Environment, and Power}{Prabhakaran and Rambow}

\firstpageno{21}

\begin{document}

\title{Dialog Structure Through the Lens of Gender,\\ Gender Environment, and Power}

\author{\name Vinodkumar Prabhakaran \email vinod@cs.stanford.edu\\
       \addr Stanford University\\
	   Stanford, CA
       \AND
       \name Owen Rambow \email rambow@ccls.columbia.edu \\
       \addr Columbia University\\
       New York, NY
       }

\editor{Raquel Fern\'{a}ndez}
\submitted{11/2016}{04/2017}{05/2017}

\maketitle

\begin{abstract}%

Understanding how the social context of an interaction affects our dialog behavior is of great interest to social scientists who study human behavior, as well as to computer scientists who build automatic methods to infer those social contexts.
In this paper, we study the interaction of power, gender, and dialog behavior in organizational interactions. 
In order to perform this study, we first construct the Gender Identified Enron Corpus of emails, in which we semi-automatically assign the gender of around 23,000 individuals who authored around 97,000 email messages in the Enron corpus. 
This corpus, which is made freely available, is orders of magnitude larger than previously existing gender identified corpora in the email domain.
Next, we use this corpus to perform a large-scale data-oriented study of the interplay of gender and manifestations of power. 
We argue that, in addition to one's own gender, the ``gender environment'' of an interaction, i.e., the gender makeup of one's interlocutors, also affects the way power is manifested in dialog. 
We focus especially on manifestations of power in the dialog structure ---  both, in a shallow sense that disregards the textual content of messages (e.g., how often do the participants contribute, how often do they get replies etc.), as well as the structure that is expressed within the textual content (e.g., who issues requests and how are they made, whose requests get responses etc.).
We find that both gender and gender environment affect the ways power is manifested in dialog, resulting in patterns that reveal the underlying factors. 
Finally, we show the utility of gender information in the problem of automatically predicting the direction of power between pairs of participants in email interactions.

\end{abstract}


\section{Introduction}

It has long been observed that men and women communicate differently in different contexts. There has been an array of studies in sociolinguistics that analyze the interplay between gender and power.
These sociolinguistic studies often rely on case studies or surveys. 
The availability of large corpora of naturally occurring interactions, and of advanced computational techniques to process the language and dialog structure of these interactions, has given us the opportunity to study the interplay between gender, power, and language use at a scale that was not feasible before. 
In this paper, we
study how gender correlates with manifestations of power in an organizational setting using the Enron email corpus.
We investigate three factors that affect choices in communication: the writer's gender, the gender of his or her fellow discourse participants
(what we call the ``gender environment''), and the power relations he or she has to the discourse participants.  
We focus on modeling the writer's choices related to discourse structure, rather than
lexical choice.
Specifically, our goal
is to show that gender, gender environment, and power all affect
individuals' choices in complex ways, resulting in patterns in the discourse that reveal the underlying factors.

We make 
three major contributions in this paper.  First, we introduce an
extension to the Enron corpus of emails: we semi-automatically
identify the
sender's gender of 87\% of email messages in the corpus.  
This extension has been made publicly available.\footnote{\url{http://www.cs.stanford.edu/~vinod/giec.html} (originally described in \cite{prabhakaran-reid-rambow:2014:EMNLP2014})} Second, 
we use this
enriched version of the corpus to investigate the interaction of hierarchical power and gender.  We formalize the notion of
``gender environment'', which reflects the gender makeup of the discourse participants of a particular conversation.  
We study how gender, power, and gender environment influence discourse participants' choices in dialog.  
This contribution 
shows how social science can benefit from advanced natural language processing techniques in analyzing corpora, allowing social scientists to tackle corpora 
that
cannot be examined in their entirety manually.  
Third, we show that the gender information in the enriched corpus can be useful for computational tasks, specifically for improving the performance of the power prediction system from our prior work \cite{prabhakaran-rambow:2014:P14-2} that is trained to
predict the direction of hierarchical power between participants in an interaction.
Our use of the gender-based features boosts 
the accuracy of predicting the direction of power between pairs of email interactants from 
68.9\% to 70.2\% on an unseen test set.

We start by discussing related work in sociolinguistics on the interplay between gender and power 
followed by work within the NLP community on gender and use of language.
In Section~\ref{sec:gender_giec}, we present the first contribution of this paper --- the Gender Identified Enron Corpus, and describe the procedure followed to build this resource and present various corpus statistics.
Section~\ref{sec:gender_genderenv} introduces the notion of gender environment and Section~\ref{sec:analframework} presents the analysis framework used in this paper.
In Section~\ref{sec:gender_stat_genderpower} and Section~\ref{sec:gender_stat_genderenvpower}, we present the statistical analysis of the interplay between gender, gender environment, and power, through the lens of dialog behavior. In Section~\ref{sec:gender_powerpred}, we demonstrate the utility of gender-based features in automatically predicting the direction of power between participants of an interaction, before we summarize our contributions in Section~\ref{sec:gender_conclusion}.

\section{Literature Review}
\label{sec:gender_related_socio}

There is much work in sociolinguistics on how gender and language use are interrelated \cite{tannen1991you,tannen1993gender,holmes1995women,kendall1997gender,coates1998language,eckert2003language,holmes2003feminine,mills2003gender,kendall200326,herring20089}.
Some of this work looks specifically at language use in work environment and/or with respect to power relations,
whereas some others study the gender differences in language use in general.
Understanding these different strands of research is important for a computational linguist working in this area. 
In this section, we summarize this literature, 
focusing more on the studies that have influenced the work presented in this paper.

\subsection{Gendered Differences in Language Use}

Many sociolinguistics studies have found evidence that men and women differ considerably in the way they communicate.
Some researchers attribute this to psychological differences \cite{gilligan1982different,boe1987language}, whereas some others suggest socialization and gendered power structures within the society as its reasons \cite{zimmerman1975sex,west1987doing,tannen1991you}. 
For instance, \newcite{tannen1991you} argues that
``for most women, the language of conversation is primarily a language of rapport: a way of establishing connections and negotiating relationships'', which she calls \textit{rapport-talk}, whereas ``for most men, talk is primarily a means to preserve independence and negotiate and maintain status in a hierarchical social order'', which she calls \textit{report-talk}.
Along the same lines, \newcite{holmes1995women} argues that ``women are much more likely than men to express positive politeness or friendliness in the way they use language''. 
In addition to politeness, many other linguistic variables have been analyzed in this context. 
\newcite{lakoff1973language} describes women's speaking style as tentative and unassertive, and argues that women use question tags and hedges more frequently than men do.
However, 
\citet{holmes1992introduction} found that the differential use of question tags in-fact depends on the function of the question tag in the interaction. She categorized the instances of question tags in terms of their functionality in the contexts in which they were used, and found that question tags used as a way to express uncertainty was done more by men, whereas question tags used as a way to facilitate communication was done more by women.
Researchers have also looked into interruption patterns in interactions in relation to gender. 
For example, \citet{zimmerman1975sex} found that men interrupted conversations more often in cross-sex interactions, whereas there were no significant differences in interruptions in same-sex interactions.

However, recent studies have suggested the need for a more nuanced view on the interplay between gender and language use.
They argue that the differences observed by above studies are due to more complex processes at play than gender alone, and that one needs to take into account the context in which the interactions happened to understand the gender differences better.
\newcite{mills2003gender} challenged the above line of analysis, especially \newcite{holmes1995women}'s theory
regarding women being more polite. 
She argues that politeness cannot be codified in terms of linguistic form alone
and calls for ``a more contextualized form of analysis, reflecting the complexity of both gender and politeness, and also the complex relation between them''.
Along those lines, \newcite{coates2013women} also challenge \newcite{lakoff1973language}'s theory on women's language being unassertive. She points out that hedges are multi-functional constructs and the greater usage of hedges by women ``can be explained in part by topic choice, in part by women's tendency to self-disclose and in part by women's preference for open discussion and a collaborative floor''.
In other words, she argues that women using more hedges than men does not entail that women are unassertive, but instead is an artifact of what topics women often take part in.
\newcite{kunsmann2013gender} connects the gender differences in language specifically to status, dominance and power. He argues that ``gender and status rather than gender or status will be the determinant categories'' of language use. In our work, we follow a similar approach. We do not study gender in isolation, but in the context of the social power relations as well as the gender environment of the interaction.

\subsection{Gender and Power in Work Place}

Within the area of studying gender and language use, there is substantial amount of work that is specifically related to the language use in work environment
\cite{west1990not,tannen1994talking,kendall1997gender,kendall200326}, mostly done through qualitative case studies.
In general, these studies found that women use more polite language and are ``less likely to use linguistic strategies that would make their authority more visible'' \cite{kendall200326}.
For instance, \newcite{west1990not} found that male physicians and female physicians differed in how they gave directives to their patients. Male physicians aggravated their directives, whereas female physicians used forms that mitigated them. 
Similarly, in the study of gender, power and language in large corporate work environments, \newcite{tannen1994talking} found that female managers use more face saving strategies (e.g., phrasing directives as suggestions: \textit{You might put in parentheses}) when talking to subordinates, whereas male managers used language that reinforced status differences (e.g., \textit{Oh, that's too dry. You have to make it snappier!}).
\newcite{kendall200326} shows that this behavior is specific to women operating in work environments. She studied the demeanor of a woman exercising her authority at work and at home, and found that while the woman used mitigating strategies to exercise her authority at work (as found by other studies before), she created a demeanor of explicit authority when exercising her authority over her daughter at home.

In this paper, we study this aspect using our formulation of overt displays of power, which are face-threatening acts that reinforce the status differences.
Our findings on the Enron emails are also in line with the above findings; we observe that male managers use significantly more overt displays of power when interacting with subordinates, whereas female managers use significantly fewer of them.
However, in contrast, we draw from a much larger-scale study in which we analyze thousands of email interactions rather than a handful of case studies in the above mentioned research.

Another line of work that has influenced our work is by \newcite{holmes2003feminine} studying the effects of gendered work environments in the manifestations of power.
They provide two case studies that analyze not the differences between male and female managers' communication, but the differences between female managers' communication in more heavily female vs. more heavily male environments. They find that, while female managers tend to break many stereotypes of ``feminine'' communication, they have different strategies in connecting with employees and exhibiting power in the two gender environments. This work has inspired us to look at this phenomenon by
formulating the notion of ``Gender Environment'' in our study. 
We adapt this notion to the level of an interaction, and define the gender environment of an email thread in terms of the ratios of
males to females on a thread, allowing us to look at whether the manifestations of power change within a more heavily male or female thread.

\subsection{Computational Approaches towards Gender and Power}

Within the NLP community, there is a considerable amount of work on
analyzing language use in relation to gender. 
Early work attempted to use NLP techniques to automatically predict the gender of authors using lexical features.
Researchers have attempted gender prediction on a variety of genres of interactions such as emails, blogs, and online social networking websites such as Twitter \cite{corney2002gender,peersman2011predicting,Cheng:2011:AGI:2296099.2296158,deitrick2012author,alowibdi2013language,nguyen-EtAl:2014:Coling}.
In more recent work, \citet{dirk2015acl} argues for research in the other direction, showing the importance of using gender information for better performance on NLP tasks such as topic identification, sentiment analysis and author attribute identification.

While automatically detecting gender is an interesting problem, our focus in this paper is not gender detection, but understanding the variations in linguistic patterns with respect to both gender and power.
For this, we require a more reliable source of gender assignments. Hence, we use publicly available name databases to reliably determine the gender of participants as we have access to the email authors' names in our corpus.
We believe that
the gender-identified email corpus we present will aid further research in the area of gender detection. Existing work on gender prediction relies on relatively smaller datasets. For example, \newcite{corney2002gender} use around 4K emails from 325 gender identified authors in their study. \newcite{Cheng:2011:AGI:2296099.2296158} use around 9K emails from 108 gender identified authors.
\newcite{deitrick2012author} use around 18K emails from 144 gender identified authors.
In contrast, we build a gender-assigned email dataset that is orders of magnitude larger than these resources. Our corpus contains around 97K emails whose authors are gender-identified, and these emails are from around 23K unique authors.

There has also been work on using NLP techniques to analyze gender differences in language use by men versus women \cite{mohammad-yang:2011:WASSA2011,bamman2012genderintwitter,bamman2014,agarwal-EtAl:2015:NAACL-HLT}.
\newcite{mohammad-yang:2011:WASSA2011} analyze the way gender affects the
expression of emotions in the Enron corpus.
They found that women send and receive emails with relatively more words that denote joy and sadness, whereas men send and receive relatively more words that denote trust and fear.
For their study, they assigned gender for the core employees in the corpus based on whether the first name of the person is easily gender identifiable or not. 
If the person had an unfamiliar name or a name that could be of either gender, they marked his/her gender as \textit{unknown} and excluded them from their study.
For example, the gender of the employee Kay Mann was marked as \textit{unknown} in their gender assignment. However, in our work, we manually research and determine the gender of
every core employee.

\citet{bamman2012genderintwitter,bamman2014}
study gender differences in the microblog site Twitter.
One of the many insights from their work is that gendered linguistic
behavior is determined by a number of factors, one of which includes the
speaker's audience, which is similar to our notion of gender environment. Their work looks at Twitter users whose linguistic style
fails to identify their gender in classification experiments, and finds
that the linguistic gender norms can be influenced by the style of their
interlocutors.
More specifically, 
people with many same-gender friends tend to use language that is strongly associated with
their gender, whereas people with more balanced social networks tend not to. 
Our notion of gender environment captures the gender makeup of an interaction, and our findings reaffirms the need to also look into the audience's gender makeup in studying gender.

NLP approaches have also been applied recently to analyzing manifestations of power in social interactions. While early studies focus on hierarchical power relations \cite{bramsen-EtAl:2011:ACL-HLT2011,Gilbert_2012,Danescu2012}, other forms of power such as situational power and influence \cite{prabhakaranCOLING2012-long,prabhakaran-rambow:2013:IJCNLP,biran2012detecting,rosenthal2014detecting,Rosenthal:2017:DIM:3068849.3014164}, power of confidence in political discourse \cite{prabhakaran-john-seligmann:2013:IJCNLP}, and pursuit of power in online forums \cite{SwayamdiptaR12} have also been explored. In \cite{prabhakaran_phdthesis}, we present a comprehensive survey of literature in this area.

To our knowledge, ours is the first computational study of this scale that focus on the interplay between gender and power in organizational email. 
We study the effects of gender in workplace interactions, not by considering the email senders' gender in isolation, but together with their power relations with the rest of the participants, as well as the gender makeup of the interaction.

\section{Gender Identified Enron Corpus}
\label{sec:gender_giec}

\label{sec:gender_data}
In this section, our starting point is the corpus (\EnronAll) used in our prior work \cite{prabhakaran-rambow:2014:P14-2}.
This corpus is derived from the Enron email corpus \cite{klimt2004enron} that contains emails from the mailboxes of 145 ``core'' Enron employees that were publicly released by the Federal Energy Regulatory Commission during its investigation of irregularities in Enron.
Our version of the corpus captures the hierarchical power relations between 13,724 pairs of employees assigned by \newcite{agarwal-EtAl:2012}, as well as the thread structure of email messages semi-automatically assigned by \newcite{Yeh06emailthread}. The thread structure allows us to go beyond isolated messages and study gender in relation to the dialog structure as well as the language use.
However, there are 34,156 unique discourse participants (senders and recipients together) across all the email threads in the corpus, and manually determining the gender of all of them is not feasible.  
Hence, we adopt a two-step approach through which we reliably
identify the gender of a large majority of discourse participants in the corpus.

\begin{enumerate}[label={Step \theenumi:},leftmargin=*]

\item Manually determine the gender of the 145 core
employees who have a bigger representation in the corpus

\item Systemically determine the gender of the rest of the discourse participants using
the Social Security Administration's baby names database

\end{enumerate}

\noindent We adopt a conservative approach so that we assign a gender only when the name of the participant
meets a very low ambiguity threshold.

\subsection{Manual Gender Assignment}
\label{sec:gender_manualgenderassignment}

We researched each of the 145 core employees using web search and found
public records about them or articles referring to them.  In order to make sure that
the results are about the same person we want, we added the word \textit{enron} to
the search queries.  Within the public records returned for each core
employee, we looked for instances in which they were being referred to either
using a gender revealing pronoun ({\em he}/{\em him}/{\em his} vs. {\em
she}/{\em her}) or using a gender revealing addressing form ({\em Mr.}
vs. {\em Mrs.}/{\em Ms.}/{\em Miss}).  Since these employees held
top managerial positions within Enron at the time of bankruptcy, it was
fairly easy to find public records or articles referring to them.  
For example, the sentence ``Kay Mann is a strong addition to Noble's senior leadership team, and we're delighted to welcome \textit{her} aboard'' (gender-revealing pronoun emphasized) in the page we found for Kay Mann clearly identifies her gender.\footnote{http://www.prnewswire.com/news-releases/kay-mann-joins-noble-as-general-counsel-57073687.html}
We were able to correctly determine the gender of each of the 145 core employees in
this manner.  A benefit of manually determining the gender of these
core employees is that it ensures a high coverage of 100\% confident gender
assignments in the corpus, as they are involved in all threads in the corpus.

\subsection{Automatic Gender Assignment}
\label{sec:gender_autogenderassignment}

\begin{figure}
\centering
\captionsetup{justification=centering}
\includegraphics[width=.6\textwidth]{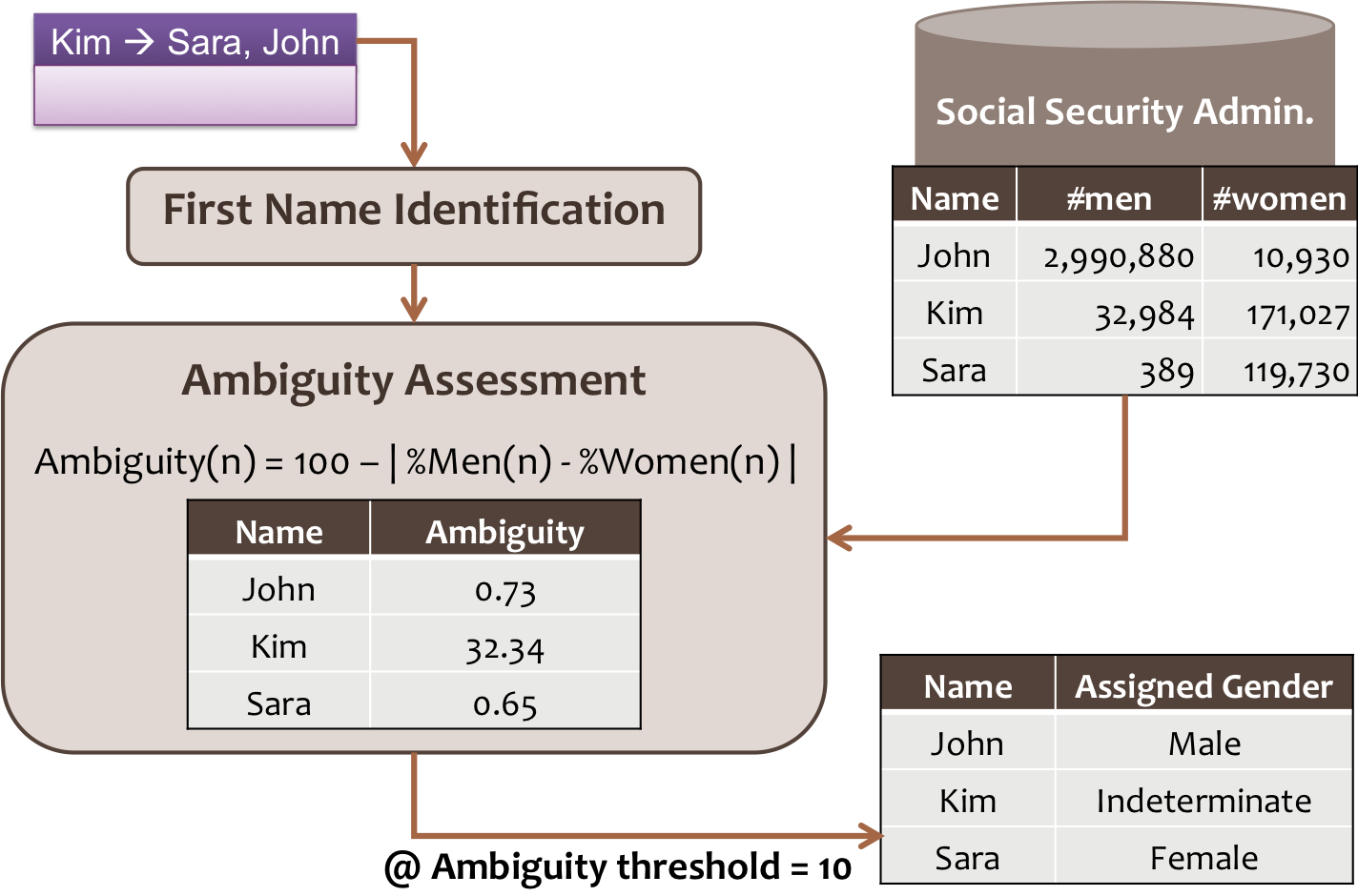}
\caption[Automatic gender assignment process]{Automatic gender assignment process. \label{fig:gender_assignment_steps}}
\end{figure}

Our corpus contains a large number of discourse participants in addition to the 145 core employees for which we manually identified the gender.  
The steps we follow to assign gender for these other discourse participants is represented graphically in Figure~\ref{fig:gender_assignment_steps}.
We first determine the first names of discourse participants and then find how ambiguous the names are by querying the Social Security Administration's (SSA) baby names dataset.  In this section, we start by describing how we calculate an ambiguity score for a name using the SSA dataset and then describe how we use it to determine the gender of discourse participants in our corpus.

\subsubsection{SSA Names and Gender Dataset}
\label{ambiguity}
 
The US Social Security Administration maintains a dataset of baby names,
gender, and name count for each year starting from the 1880s, for names with at
least five
counts.\footnote{http://www.ssa.gov/oact/babynames/limits.html} We used
this dataset in order to determine the gender ambiguity of a name.  The
Enron data set contains emails from 1998 to 2001. 
We estimate the common age range for
a large, corporate firm like Enron
at 24-67,\footnote{http://www.bls.gov/cps/demographics.htm} so we used the
SSA data from 1931-1977 to calculate ambiguity scores for our purposes.

For each name $ n $ in the database, let $ \mathit{mp}(n) $ and $ \mathit{fp}(n)
$
denote the percentages of males and females with the name $ n $. 
The difference between these percentages of a name gives us a measure of how ambiguous it is; the smaller the difference, the more ambiguous the name.
We define the ambiguity score of a name $ n $, denoted by $ \mathit{AS}(n) $, as follows:
\[
\mathit{AS}(n) = 100 - |\mathit{mp}(n) - \mathit{fp}(n)| 
\]
\noindent The value of $ \mathit{AS}(n) $ varies between 0 and 100. 
A name that is `perfectly unambiguous' would have an ambiguity score of 0, while a `perfectly ambiguous' name (i.e., 50\%/50\% split between genders) would have an ambiguity score of 100.
We assign the likely gender of the name to be the one with the higher
percentage, if the ambiguity score is below a threshold $ \mathit{AS}_T $.

\[
G(n) = 
\begin{cases}
    Male (M), & \text{if } \mathit{AS}(n) \le \mathit{AS}_T \text{ and } mp(n) > fp(n) \\
    Female (F), & \text{if } \mathit{AS}(n) \le \mathit{AS}_T \text{ and } mp(n) < fp(n) \\
    Indeterminate (I), & \text{if }  \mathit{AS}(n) > \mathit{AS}_T
\end{cases}
\]

\begin{figure}[t]
\centering
\captionsetup{justification=centering}
\includegraphics[width=.5\textwidth]{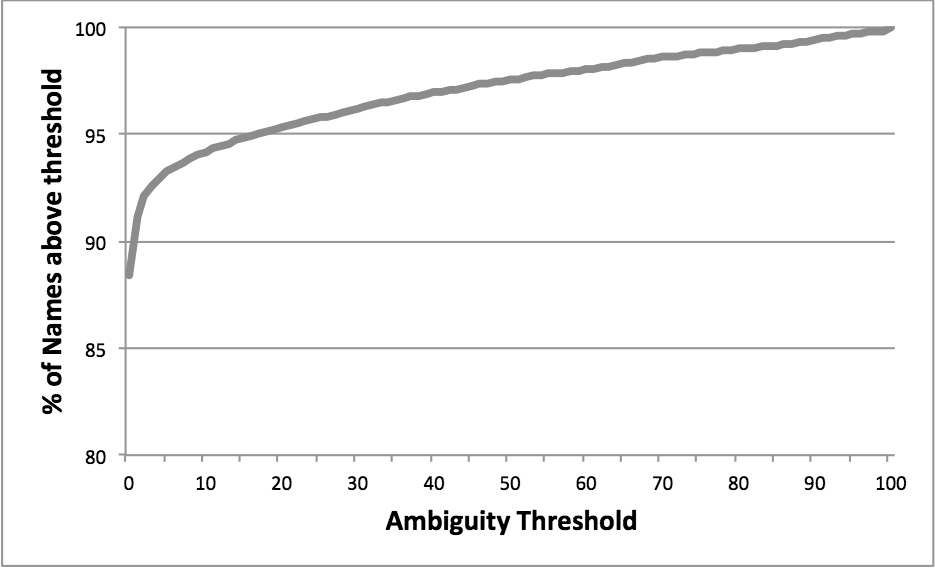}
\caption[Plot of percentage of first names covered against ambiguity threshold]{Plot of percentage of first names covered against ambiguity threshold. \label{fig:gender_ambiguityplot}}
\end{figure}

Figure~\ref{fig:gender_ambiguityplot}
shows the plot of the percentage of names that will be gender assigned in the SSA dataset against the ambiguity threshold.
As the plot shows, around 88\% of the names in the SSA dataset have $ \mathit{AS}(n) = 0$, i.e., are unambiguous. 
We choose a very conservative threshold of $ \mathit{AS}_T = 10 $ for our gender assignments, which assigns gender to around 93\% names in the SSA dataset. 
An ambiguity threshold of $ 10 $ means that we assign a gender only if at least 95\% of people with that name were of that gender. 
In the gender assigned corpus that we released, we retain the $ \mathit{AS}(n) $ of each name, so that the users of this resource can decide the threshold that suits their needs.

\subsubsection{Identifying the First Name}

Each discourse participant in our corpus has at least one email address and
zero or more names associated with it. The name field is automatically
assembled by \newcite{Yeh06emailthread}, who captured the different names
from email headers. The names in the email headers are populated from individual email clients the senders were using and hence do not follow a standard format. 
To make things worse, not all discourse participants are human; some may refer to organizational groups (e.g., HR Department) or anonymous corporate email accounts (e.g., a webmaster
account, do-not-reply address etc.). 
The name field may sometimes be empty, contain multiple names, contain an email address, or show other
irregularities.  Hence, it is nontrivial to determine the first name of
our discourse participants.  We used the heuristics below to extract
the set of candidate names for each discourse participant.

\begin{itemize}
\item If the name field contains two words, 
pick the second or first word, depending on whether a comma separates them or not; pick the first word if the name field does not contain a comma; pick the word following the comma if it does contain one.
\item If the name field contains three words and a comma, choose the second and third words (a likely first and middle name, respectively). If the name field contains three words but no comma, choose the first and second words (again, a likely first and middle name).
\item If the name field contains an email address, pick the portion from the beginning of the string to a `.',`\_' or `-'; if the email address is in camel case, take portion from the beginning of the string to the first upper case letter.
\item If the name field is empty, apply the above rule to the email address field to pick a name.
\end{itemize}

\noindent In addition, we cleaned up some irregularities that were present in the name field. One common issue was that many email fields started with the text ``?S" possibly a manifestation of some data preprocessing step. We strip this portion of the string in order to obtain the part that denote the actual email address.

The above heuristics create a list of candidate names for each discourse participant. For each candidate name, we compute the ambiguity score (Section~\ref{ambiguity}) and the likely gender. 
We find the candidate name with the lowest ambiguity score that passes the threshold and assign the associated gender to the discourse participant. 
If none of the candidate names for a discourse participant passes the threshold,
we assign the gender to be indeterminate.
We also assign the gender to be indeterminate, if none of the candidate names is present in the SSA dataset.
This will occur if the name is a first name that is not in the database (an unusual or international name; e.g., \textit{Vladi}), or if no true first name was found (e.g., the name field
was empty and the email address was only a pseudonym). This will also include most of the cases
where the discourse participant is not a human (e.g., \textit{HR Department}).

 \subsubsection{Coverage and Accuracy}
 
We evaluated the coverage and accuracy of our gender assignment system on the
manually assigned gender data of the 145 core people. 
We obtained a
coverage of 90.3\%, i.e., for 14 of the 145 core people, either their name's ambiguity
score was higher than the threshold 
(\textit{Kam},
\textit{Lindy},
\textit{Tracy},
\textit{Lynn},
\textit{Chris},
\textit{Stacy},
\textit{Robin},
\textit{Stacey}, and
\textit{Tori})
or their name did not exist in the SSA dataset (\textit{Geir} and \textit{Vladi}).  Of the 131 people the system assigned
a gender to, we obtained an accuracy of 89.3\% in correctly identifying the gender. We investigated the errors
and found that all errors were caused due to incorrectly identifying
the first name.  
For the cases where we correctly identify the first name, we obtain a 100\% accuracy in assigning the gender.
The errors in finding first name arise because the name fields are automatically populated and 
sometimes the core discourse participants' name fields include their
secretaries' who are of the other gender.
While the name fields capturing multiple people is common for people in higher managerial positions, we expect this not to happen in the middle 
management and below, to which most of the automatically gender-assigned discourse participants belong.

\subsection{Corpus Statistics and Divisions}
\label{sec:gender_corpusstats}

\paragraph{Gender assignment coverage:}
We apply the gender assignment system described above to all discourse
participants of all email threads in the \EnronAll corpus to build the Gender Identified Enron Corpus (GIEC). 
Table~\ref{table:gendercorpstats1} shows the coverage of gender assignment in the GIEC corpus at different levels: unique discourse participants, messages and threads.
We were able to identify the gender of 67\% of unique discourse participants in the corpus. 
We verified that a majority of the cases where we could not assign the gender was due to the name of the sender email account not being present in the SSA dataset --- mostly, cases where the discourse participant is not a human (e.g., \textit{HR Department}) as well as one-off email addresses (without a name entry) from outside the Enron.
In fact, the 67\% discourse participants whose gender we could identify amounted to the senders of 87\% of the messages in our corpus.
We call the subset of threads for which we were able to identify the gender of all email senders, the \textit{All Senders Gender Identified (ASGI)} sub-corpus, and those for which we were able to identify the gender of all participants including senders and all recipients, the \textit{All Participants Gender Identified (APGI)} sub-corpus.
ASGI covers around 71\% of threads in the corpus, whereas APGI covers only about 49\%.
The users of this resource can limit their study to either subset, depending on their requirements.

\begin{figure*}[ht]
	\centering
	\captionsetup{justification=centering}
	\begin{subfigure}[t]{0.49\textwidth}
		\centering
		\includegraphics[width=\textwidth]{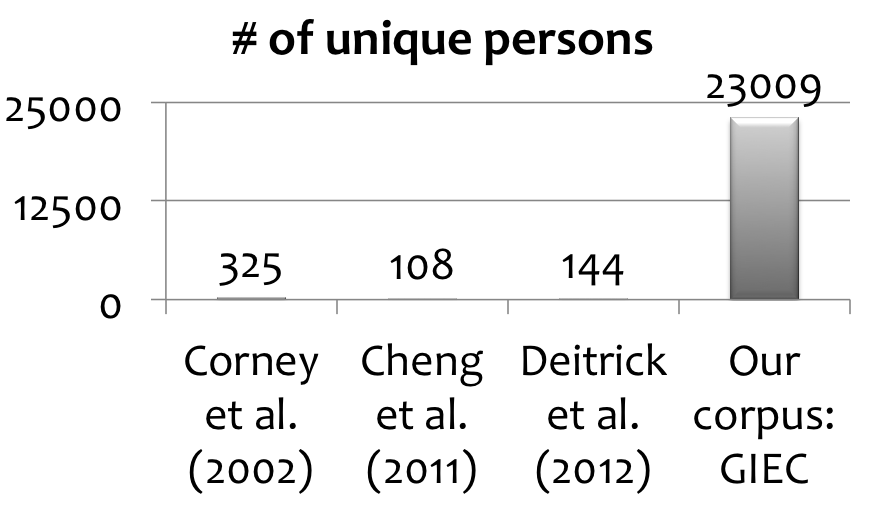}
		\caption{Comparison in terms of number of unique discourse participants}
	\end{subfigure}%
	~ 
	\begin{subfigure}[t]{0.49\textwidth}
		\centering
		\includegraphics[width=\textwidth]{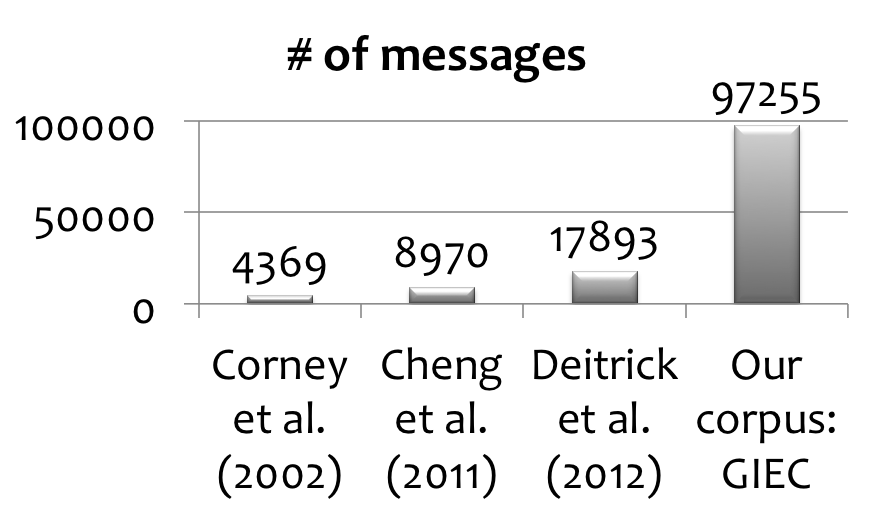}
		\caption{Comparison in terms of number of messages}
	\end{subfigure}
	\caption[Gender Identified Enron Corpus vs. existing gender assigned resources]{\label{figure:giec_stat_comparison} Gender Identified Enron Corpus (GIEC) vs. existing gender assigned resources.}
\end{figure*}

In Figure~\ref{figure:giec_stat_comparison}, we show how the size of our Gender Identified Enron Corpus compares to existing gender assigned corpora within the emails domain \cite{corney2002gender,Cheng:2011:AGI:2296099.2296158,deitrick2012author}. Our corpus is orders of magnitude larger than existing resources.
We have representation of over 23K authors in our corpus, as opposed to a few hundred in other existing resources. In terms of number of messages also, our corpus is more than 5 times the size of next biggest corpus.

\begin{table}
\centering
\captionsetup{justification=centering}
\begin{tabular}{l c }
\toprule
 & \multicolumn{1}{c}{Count (\%)} \\ 
\midrule
Total unique discourse participants & \multicolumn{1}{c}{34,156} \\
- gender identified  & 23,009 (67.3\%) \\
\hline
Total messages & \multicolumn{1}{c}{111,933} \\
- senders gender identified  & 97,255 (86.9\%) \\
\hline
Total threads & \multicolumn{1}{c}{36,615} \\
- All Senders Gender Identified (ASGI) & 26,015 (71.1\%) \\
- All Participants Gender Identified (APGI) & 18,030 (49.2\%) \\
\bottomrule
\end{tabular}
\caption[Coverage of gender identification at various levels]{\label{table:gendercorpstats1} Coverage of gender identification at various levels: unique discourse participants, messages and threads.}
\end{table}

\begin{figure}[h]
\centering
\captionsetup{justification=centering}
\includegraphics[width=\textwidth]{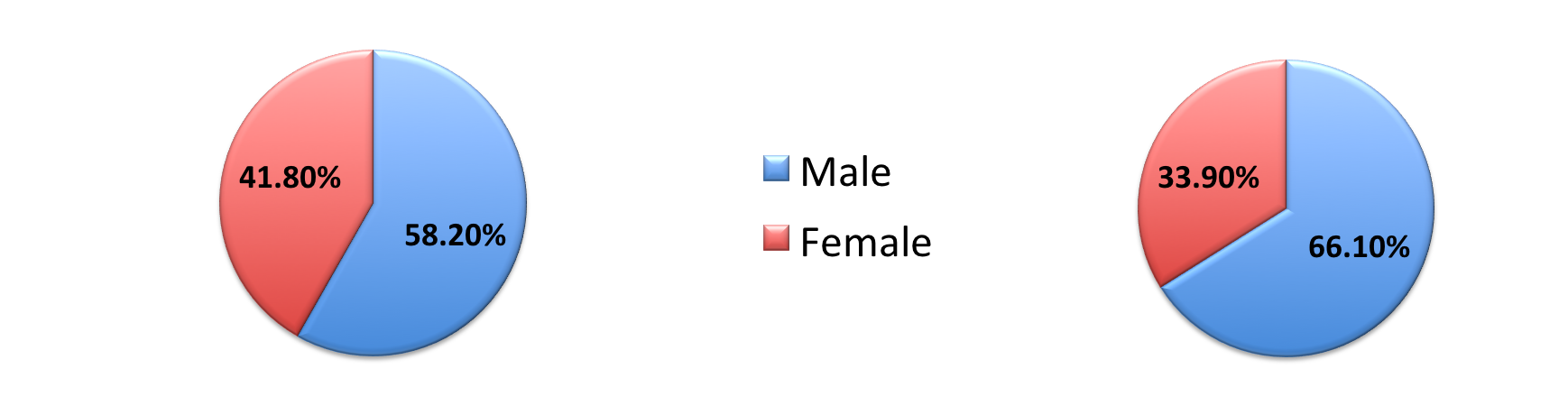}
\caption[Male/Female split in gender assignments]{\label{figure:automatic_gender_assignment_split} Male/Female split in gender assignments across a) all unique participants who were gender identified (left), b) all messages whose senders were gender identified (right)}
\end{figure}

\paragraph{Gender assignment male/female split:}
In Figure~\ref{figure:automatic_gender_assignment_split}, we show the male/female percentage split of all unique discourse participants, as well as the split at the level of messages (i.e., messages sent by males vs. females).
We have more male participants than female participants in the corpus (58\% vs. 42\%). When counted in terms of number of messages, around two thirds of the messages in our corpus were sent by men.

\section{Notion of Gender Environment}
\label{sec:gender_genderenv}

In this study, we are interested not only in how the gender of a discourse participant affects their dialog behavior, but also whether the genders of other participants they are interacting with has an effect on their dialog behavior.
We use the term ``gender environment'' to refer to the gender composition of a
group who are communicating.  
We derive this notion from
\citet{holmes2003feminine} in which  the term is used to refer
to a stable work group who
interact regularly.  Since we are interested in studying email conversations
(threads), we adapt this notion to refer to a single thread at a time.
We consider the ``gender
environment'' to be specific to each discourse participant and to describe
the other participants from his or her point of view.  Put differently, we
use the notion of ``gender environment'' to model a discourse participant's
(potential) audience in a conversation.  For example, a conversation among
five women and one man looks like an all-female audience from the man's
point of view, but a majority-female audience from the women's points of
view.

We define the gender environment of a discourse participant $p$ in a thread
$t$ as follows.
As discussed, we assume that the gender environment is a property of each
discourse participant $p$ in thread $t$.  
We take the set of all discourse
participants of the thread $t$, $P_t$, and
exclude $p$ from it: $P_t \setminus \{p\}$.  We then calculate the
percentage of females in this set.\footnote{We note that one could also define the notion of gender environment at the level of individual emails: not all emails in a thread involve the same set of participants. We leave this to future work.}  
We obtain three gender environments by setting
thresholds on these percentages (dividing equally): Female Environment, Mixed Environment, and Male Environment.

\begin{itemize}
	
	\item
	
	{\bf Female Environment}: if the percentage of women in
	$P_t \setminus \{p\}$ is above 66.7\%.
	
	\item
	
	{\bf Mixed Environment}: if the percentage of women in
	$P_t \setminus \{p\}$ is between 33.3\% and 66.7\%.   
	
	\item
	
	{\bf Male Environment}: if the percentage of women in
	$P_t \setminus \{p\}$ is below 33.3\%

\end{itemize}

\section{Analysis Framework}
\label{sec:analframework}

In the rest of this paper, we use the All Participants Gender Identified (APGI) subset of the Enron corpus to study the interplay of gender and power, as it allows us to study the effects of both Gender and Gender Environment. 
We use the same analysis framework --- problem formulation, data splits, and features --- introduced in \cite{prabhakaran-rambow:2014:P14-2}.
In this section, we briefly summarize the analysis framework and features we used. For a detailed account of the problem and features, refer to \cite{prabhakaran-rambow:2014:P14-2}.

\subsection{Power Annotations}

Our corpus contains organizational hierarchy relations extracted by \newcite{agarwal-EtAl:2012} from the Enron organizational charts. 
They define a dominance relation to be the relation between superior and subordinate in the hierarchy. Their gold standard for hierarchy relations contains a total of 1,518 employees. They
found 2,155 immediate dominance relations spread over 65 levels of dominance (CEO, manager,
trader etc.) among these 1,518 employees. They also added the transitive closure of these relations to the corpus resulting in a total of 13,724 dominance relations. We use these dominance relations as our gold standard for assigning superior-subordinate relations.

\subsection{Problem Formulation}

Let $ \mathit{t} $ denote an email thread and $ \mathit{M_t} $ denote the set of all messages in $ \mathit{t} $. Also, let $ \mathit{P_t} $ be the set of all participants in $ \mathit{t} $, i.e., the union of senders and recipients (\textit{To} and \textit{CC}) of all messages in $ \mathit{M_t} $.
We are interested in detecting power relations between pairs of participants who interact within a given email thread.
Not every pair of participants $ \mathit{(p_1, p_2) \in P_t \times P_t}$ interact with one another within $ \mathit{t} $. 
Let $ \mathit{IM_t(p_1, p_2)}$ denote the set of \textit{Interaction Messages} --- non-empty messages in $ t $ in which either $ p_1 $ is the sender and $ \mathit{p_2} $ is one of the recipients or vice versa.
We call  
the set of $ \mathit{(p_1, p_2)} $ such that $ \mathit{|IM_t(p_1, p_2)| > 0} $
 the \textit{interacting participant pairs} of $ \mathit{t} $ ($\mathit{IPP_t}$). 
We focus on the 
manifestations of power in interactions between people 
across
different levels of hierarchy.
For every $ \mathit{(p_1, p_2) \in IPP_t}$, 
we query the set of dominance relations in the gold hierarchy 
to determine their hierarchical power relation ($\mathit{HP(p_1, p_2)}$). 
We exclude pairs that do not exist in the gold hierarchy from our analysis and
denote the remaining set of 
\textit{related interacting participant pairs} as $\mathit{RIPP_t}$.
We assign $\mathit{HP(p_1, p_2)}$ to be
\textit{superior} if $ \mathit{p_1} $ dominates $ \mathit{p_2} $, and \textit{subordinate} if $ \mathit{p_2} $ dominates $ \mathit{p_1} $.
In this paper, we are interested in how gender interacts with the differences in dialog behavior exhibited by superiors and subordinates. We study how a participant's gender and the gender of other participants in an email thread affects these dialog behavior differences.

We formulate the problem as a computational task. 
Given a thread $ \mathit{t} $ and a pair of participants $ \mathit{(p_1, p_2) \in RIPP_t}$, we want to automatically detect $\mathit{HP(p_1, p_2)}$.
This problem formulation is similar to the ones in \cite{bramsen-EtAl:2011:ACL-HLT2011} and \cite{Gilbert_2012}.
However, the difference is that for us an instance is a pair of 
participants in a single thread of interaction
(which may or may not include other people),
whereas for them an instance constitutes all messages exchanged between a pair of people in the entire corpus.
Our formulation also differs from
\cite{prabhakaran-rambow:2013:IJCNLP} in that we detect power relations between pairs of participants,
instead of
just whether a participant had power over anyone in the thread.

\subsection{Data}

We follow the same
\textit{train}, \textit{dev}, \textit{test} division of \EnronAll as in \cite{prabhakaran-rambow:2014:P14-2}. 
We limit our study to the threads in which were able to identify the gender of all participants (i.e., threads that are part of the APGI subset of the corpus). 
Table~\ref{table:gender_enron_datastats} presents the total number of pairs in
$\mathit{IPP_t}$ and $\mathit{RIPP_t}$ from all the threads in the APGI subset of our corpus
and across the 
\textit{train}, \textit{dev} and \textit{test} sets.
We choose APGI instead of ASGI (All Senders Gender Identified) because APGI allows us to also study the notion of Gender Environment for which we need to know the gender of all participants. 
As an artifact of choosing the APGI, we also have a corpus with relatively smaller number of participants per thread than the full corpus. In other words, email threads with a large number of participants, such as broadcast emails, will have been excluded from the AGPI, since there is a higher chance that the automatic gender assignment step fails to assign the gender for at least one of the recipients.
As a result, 
the findings from the analysis on this subset sometimes differ from what we found in \cite{prabhakaran-rambow:2014:P14-2}.
However, knowing how the two corpora differ in terms of the number of participants, 
it is interesting to note on which aspects of interactions the findings in both studies differ.

\begin{table}
	\centering
	\captionsetup{justification=centering}
	\begin{tabular}{l r r r r}
		\toprule
		Description & Total & Train & Dev & Test  \\ 
		\midrule
		\# of threads  & 17,788  & 8,911 & 4,328 & 4,549 \\
		$\mathit{ \sum_t|IPP_t| } $ & 74,523 & 36,528 & 18,540 & 19,455  \\
		$\mathit{ \sum_t|RIPP_t| } $ &  4,649 & 2,260 & 1,080 & 1,309 \\
		\bottomrule
	\end{tabular}
	\caption[Data statistics in the All Participants Gender Identified subset of the Gender Identified Enron Corpus]{\label{table:gender_enron_datastats}Data statistics in the All Participants Gender Identified subset of the Enron Corpus.}
	Row 1 presents the total number of threads in different subsets of the corpus. \\
	Row 2 and 3 present the number of interacting participant pairs ($ \mathit{IPP} $) and related interacting participant pairs ($ \mathit{RIPP} $)
	in those subsets.
\end{table}

\subsection{Features}

\label{sec_features}

We study the same dialog structural aspects of interaction introduced from \cite{prabhakaran-rambow:2014:P14-2} in this work. 
In this section we briefly describe the various features we use to model these aspects of interactions.
We focus on features in five different dialog structural aspects of interactions 
---  
\featSetPositional, \featSetVerbosity, \featSetMetaData, \featSetDialogActs, and \featSetOvertDisplayOfPower, as well as a non-structural aspect captured by \featSetNgrams features. The first three aspects (\featSetPositional, \featSetVerbosity, and \featSetMetaData) capture the structure of message exchanges without doing any NLP processing on the content of the emails (e.g., how many emails did a person send), whereas
\featSetDialogActs and \featSetOvertDisplayOfPower
capture the pragmatics of the dialog and require an analysis of the content of the emails (e.g., did they issue any requests). 
\featSetNgrams features also analyze the content, but at a shallow level, looking solely at word lemma and part-of-speech ngrams.

Each feature $ \mathit{f} $ is extracted with respect to a person $ \mathit{p} $ 
over a reference set of messages $ \mathit{M} $ (denoted $ \mathit{f^p_M }$).
For example, $ \mathit{MsgRatio^{Kim}_{M_t}}$ denotes the ratio of messages sent by \textit{Kim} to the total number of messages in the thread $ t $, whereas $ \mathit{MsgRatio^{Sara}_{IM_t(Kim,Sara)}}$ denotes the ratio of messages sent by \textit{Sara} to the total number of interaction messages between \textit{Kim} and \textit{Sara} in the thread $ t $.
For each pair $ \mathit{(p_1, p_2)} $, we extract 4 versions of each feature $ \mathit{f}$.

\begin{table}[h]
	\centering
	\setlength{\extrarowheight}{.1cm}
	\begin{tabular}{l l}
		
		$ \mathit{f^{p_1}_{IM_t(p_1, p_2)}} $: & features with respect to $ p1 $ and interaction messages between $ p1 $ and $ p2 $\\ 
		$ \mathit{f^{p_2}_{IM_t(p_1, p_2)}} $: & features with respect to $ p2 $ and interaction messages between $ p1 $ and $ p2 $ \\
		$ \mathit{f^{p_1}_{M_t}} $: & features with respect to $ p1 $ and all messages in thread $ t $ \\
		$ \mathit{f^{p_2}_{M_t}} $: & features with respect to $ p2 $ and all messages in thread $ t $ \\
		
	\end{tabular}
\end{table}

\noindent The first two versions capture behavior of the pair among themselves, while the third and fourth capture their overall behavior in the entire thread.
In Table~\ref{table:gender_enron_dimensions}, we list each feature $ \mathit{f} $ we use. 
Like \cite{prabhakaran-rambow:2014:P14-2}, we use all four versions of the features in the machine learning experiments. However, for the statistical analysis presented in Section~\ref{sec:gender_stat_genderpower} and Section~\ref{sec:gender_stat_genderenvpower}, we use the $ \mathit{f^{p_1}_{M_t}} $ version alone (similar results were obtained using the $ \mathit{f^{p_1}_{IM_t(p_1, p_2)}} $ version as well).

\begin{table*}[t]
\centering

\begin{tabular*}{\textwidth}{l | l | l }

\toprule

Aspects & Features & Description \\

\midrule
\multirow{3}{*}{\featSetPositionalShort}
& \feat{Initiator} & did $ \mathit{p} $ sent the first message? \\
& \feat{FirstMsgPos} & relative position of $ \mathit{p} $'s first message in $ \mathit{M} $ \\
& \feat{LastMsgPos} & relative position of $ \mathit{p} $'s last message in $ \mathit{M} $ \\

\midrule
\multirow{5}{*}{\featSetVerbosityShort}
& \feat{MsgCount}  & Count of messages sent by $ \mathit{p} $ in $ \mathit{M} $ \\
& \feat{MsgRatio} & Ratio of messages sent in $ \mathit{M} $\\
& \feat{TokenCount} &  Count of tokens in messages sent by $ \mathit{p} $ in $ \mathit{M} $ \\
& \feat{TokenRatio} &  Ratio of tokens across all messages in $ \mathit{M} $ \\
& \feat{TokenPerMsg} &  Number of tokens per message in messages sent by $ \mathit{p} $ in $ \mathit{M} $ \\

\midrule
\multirow{6}{*}{\featSetMetaDataShort}
& \feat{AvgRecipients} & Avge. number of recipients in messages \\
& \feat{AvgToRecipients} & Avge. number of To recipients in messages \\
& \feat{InToList\%} & \% of emails $ \mathit{p} $ received in which he/she was
in the To list \\
& \feat{AddPerson} & did $ \mathit{p} $ add people to the thread? \\
& \feat{RemovePerson} & did $ \mathit{p} $ remove people to the thread? \\
& \feat{ReplyRate} & average number of replies received
per message by $ \mathit{p} $ \\

\midrule
\multirow{5}{*}{\featSetDialogActsShort}
& \feat{ReqActionCount} & \# of Request Action dialog acts in $ \mathit{p} $'s messages \\
& \feat{ReqInformCount} & \# of Request Information dialog acts in $ \mathit{p} $'s messages \\
& \feat{InformCount} & \# of Inform dialog acts in $ \mathit{p} $'s messages \\
& \feat{ConventionalCount} & \# of Conventional dialog acts in $ \mathit{p} $'s messages \\
& \feat{DanglingReq\%} & \% of $ \mathit{p} $'s messages with requests that did not have a reply \\

\midrule
\multirow{1}{*}{\featSetOvertDisplayOfPowerShort}
& \feat{ODPCount} & Number of instances of overt displays of power \\

\midrule
\multirow{3}{*}{\featSetNgramsShort}
& \feat{LemmaNGram} & Word lemma ngrams \\
& \feat{POSNGram} & Part of speech (POS) ngrams \\
& \feat{MixedNGram} & POS ngrams, with closed classes replaced with lemmas\\

\bottomrule

\end{tabular*}

\caption[Aspects of interactions analyzed in organizational emails]{\label{table:gender_enron_dimensions}Aspects of interactions analyzed in organizational emails.}

\end{table*}

\subsubsection{Positional Features}

There are three features in this category --- \featInitiator, \featFirstMsgPos, and \featLastMsgPos.
\featInitiator is a boolean feature which gets the value of 1 (\textit{true}) if the $ p $ sent the first message in the thread, and 0 otherwise (\textit{false}). \featFirstMsgPos, and \featLastMsgPos are real-valued features taking values from 0 to 1, capturing relative positions of $ \mathit{p} $'s first and last messages. The lower the value, the earlier the participant sent his/her first (or last) message. 
The first two features relate to the participant's initiative. 
\featLastMsgPos captures whether the participant stays till the end of the email thread.

\subsubsection{Verbosity Features}

This set of features captures how verbose were the participants in the thread. 
There are five features in this set --- \featMsgCount, \featMsgRatio, \featTokenCount, \featTokenRatio, and \featTokenPerMsg.
The first two features measure verbosity in terms of $ p $'s messages (raw counts and percentages), whereas the third and fourth features measure verbosity in terms of word tokens in $ p $'s messages (raw counts and percentage). The last feature measure how terse or verbose on average $ p $'s messages are.

\subsubsection{Thread Structure Features}

This set of features captures the structure of the email in terms of meta-data that is part of the email headers. It includes seven features --- 
\feat{AvgRecipients}, \feat{AvgToRecipients}, \feat{InToList\%}, \feat{AddPerson}, \feat{RemovePerson}, and \feat{ReplyRate}. 
The first two features capture the `reach' of the person in terms of the average number of total recipients as well as
recipients in the To list in emails sent by $ \mathit{p} $.
\feat{InToList\%} capture the 
the percentage of emails $ \mathit{p} $ received in which he/she was in the \textit{To} list (as opposed to the \textit{CC} list); 
The next two features ---\feat{AddPerson} and \feat{RemovePerson}--- are boolean features denoting whether $ p $ added or removed people when responding to a message.
Next, we look at the responsiveness towards $ p $ as the 
average number of replies received per message sent by $ \mathit{p} $ (\feat{ReplyRate}).

\subsubsection{Dialog Act Features}

This feature set contains features that capture the dialog acts used by participants in the thread. 
We obtain dialog act tags on the entire corpus using the automatic dialog act tagger from our previous work \cite{omuya-prabhakaran-rambow:2013:NAACL-HLT}.
The DA tagger labels each sentence to be one of the 4 dialog acts: 

\begin{itemize}
	
	\item \reltype{Request-Action}: the writer signals her desire that the
	reader 
	perform some non communicative act, i.e., an act that cannot in
	itself be part of the dialogue.  For example, a writer can ask the
	reader to write a report or make coffee.
	
	\item \reltype{Request-Information}: the writer signals her desire that
	the reader perform a specific communicative act, namely that he 
	provide information (either facts or opinion).  
	
	\item \reltype{Inform}: the writer conveys information, or
	more precisely, the writer signals her desire that the reader
	adopt a certain belief.  It covers many different types of
	information that can be conveyed including answers to questions,
	beliefs (committed or not), attitudes, and elaborations on prior
	DAs.
	
	\item \reltype{Conventional}: dialog act does not signal any
	specific communicative intention on the part of the writer, but
	rather it helps structure and thus facilitate the communication.
	Examples include greetings, introductions, expressions of
	gratitude, etc.
	
\end{itemize}
\label{sec:datagger}
\noindent The tagger uses a cascaded minority preference multi-class algorithm that posted significant improvements in its performance of identifying minority dialog acts such as Request Action (23\% error reduction over the one-vs-all classification algorithm), and
obtained an overall accuracy of 92\%.
Please refer to \cite{omuya-prabhakaran-rambow:2013:NAACL-HLT} for more details on the dialog act tagging framework.
We use 4 features: 
\feat{ReqActionCount}, \feat{ReqInformCount}, \feat{InformCount}, and \feat{ConventionalCount} 
to capture the number of sentences in messages sent by $ \mathit{p} $ that has each of these labels, respectively.
We also use a feature to capture the percentage of $ \textit{p} $'s messages that had a request (either \DAReqAction or \DAReqInform), which did not get a reply,
i.e., dangling requests (\feat{DanglingReq\%}).

\subsubsection{Overt Display of Power}

We use the notion of Overt Display of Power (ODP) introduced in our prior work \cite{prabhakaranODP2012-long} to measure face aggravating acts in the interactions.
We define an utterance to have ODP if it is interpreted as creating additional constraints
on the response beyond those imposed by the general dialog act.
For example, ``I need the report by end of Friday'' would be considered as an overt display of power, whereas ``Could you please try to send the report by end of Friday'' would not be considered as one.
We consider ODP as a pragmatic concept, i.e., in terms of the dialog constraints an utterance introduces to its response, and not in terms of specific linguistic markers. For example,
the use of politeness markers (e.g., \textit{please}) does not, on its own, determine the presence or absence of an ODP. In addition, the presence of ODP cannot be determined solely based on syntactic patterns alone (e.g., declarative sentences such as \textit{I need the report} may also function as ODPs). 

In \cite{prabhakaranODP2012-long}, we presented a data-oriented approach of identifying instances of ODPs in email threads. We first obtained manual annotations of ODP on a subset of 122 email threads (1734 sentences) at the sentence level, and then built an SVM-based supervised machine learning model to identify instances of ODP in new email threads. In addition to lexical features, it also uses the dialog act features obtained using the  dialog act tagger described in Section~\ref{sec:datagger}. Our ODP tagger has an accuracy of 96\% and an F-measure of 54\% over a random prediction baseline F-measure of 10.4\%.

In this paper, we applied the above ODP Tagger to the email threads in our entire corpus and used a feature 
\feat{ODPCount} that captures number of instances of overt displays of power in $ \mathit{p} $'s messages.

\subsubsection{Lexical Features}

In addition to the dialog structure features, we also used simple lexical ngram features as they have already been shown to be valuable in predicting power relations \cite{bramsen-EtAl:2011:ACL-HLT2011,Gilbert_2012}. 
We use the feature set \featSetNgrams to capture word lemma ngrams, POS (part of speech) ngrams and mixed ngrams. A mixed ngram is a special case of word ngram where words belonging to open classes are replaced with their POS tags, thereby being able to capture longer sequences without increasing the dimensionality as much as word ngrams do.
We found the best setting
to be using both unigrams and bigrams for all three types of ngrams, by tuning on our \textit{dev} set.

\section{Gender and Power: A Statistical Analysis}
\label{sec:gender_stat_genderpower}

As a first step, we would like to understand whether male superiors, female superiors, male subordinates, and female subordinates differ in their dialog behavior. For this analysis, the ANOVA (Analysis of Variance) test is the appropriate statistical test as it provides a way to test whether or not the means of several groups are equal. In other words, ANOVA generalizes the Student's t-Test to situations with more than two groups. It also eliminates the possibility of making a type I error (false positives) if multiple two-sample t-Tests are applied to such a problem.
We perform ANOVA tests on all dialog structure features --- \featSetPositional, \featSetVerbosity, \featSetMetaData, \featSetDialogActs, and \featSetOvertDisplayOfPower
keeping both Hierarchical Power and Gender as independent variables. This results in four groups --- male superiors, female superiors, male subordinates, and female subordinates.
It is crucial to note that ANOVA only determines that there is a significant difference between groups, but does not tell which groups are significantly different. In order to ascertain that, we use the Tukey's HSD (Honest Significant Difference) Test. 
We discuss the significant findings from these analyses below.

Altogether, there are twenty features as dependent variables, and two independent variables --- Power and Gender. That is a total of sixty different statistical tests; in addition, for each ANOVA test, we also perform the Tukey's HSD test. 
Even after applying the Bonferroni correction to control for multiple testing (i.e., significance level at 0.05/120=0.0008), many of the results we discuss below hold statistical significance. Hence, our overall hypothesis that gender affects the way power is manifested in interactions holds true. However, as an exploratory study, we present the results along each individual aspect without applying the correction, as it has been shown that the Bonferroni correction tends to be conservative. 

\subsection{Positional Features}
\label{sec:gender_positional_results}
\label{sec:power_gender_pst}

There are three features in this category --- \featInitiator, \featFirstMsgPos, and \featLastMsgPos.
\featInitiator is a binary feature which gets the value of 1 (\textit{true}) if the participant sent the first message in the thread, and 0 otherwise (\textit{false}). 
\featFirstMsgPos and \featLastMsgPos are real-valued features taking values from 0 to 1. The lower the value, the earlier the participant sent the first (or last) message. 
The first two features relate to the participant's initiative. A higher average value for \featInitiator in a group indicates that participants in that group initiates threads more often; so does a lower average value for \featFirstMsgPos.
\featLastMsgPos captures whether participant stayed on towards the end of the thread.

\begin{figure*}
	\centering
	\includegraphics[width=0.4\linewidth]{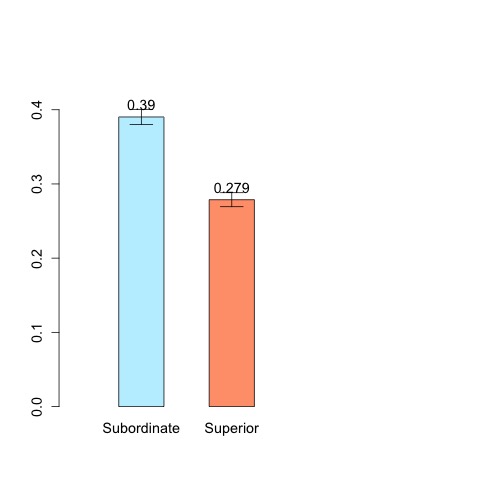}
	\hspace{-1.8cm}
	\includegraphics[width=0.4\linewidth]{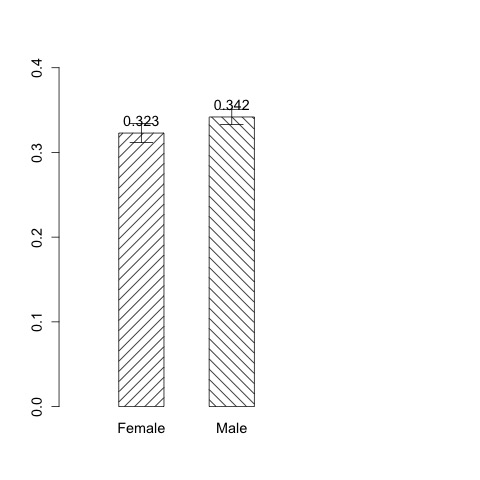}
	\hspace{-1.8cm}
	\includegraphics[width=0.4\linewidth]{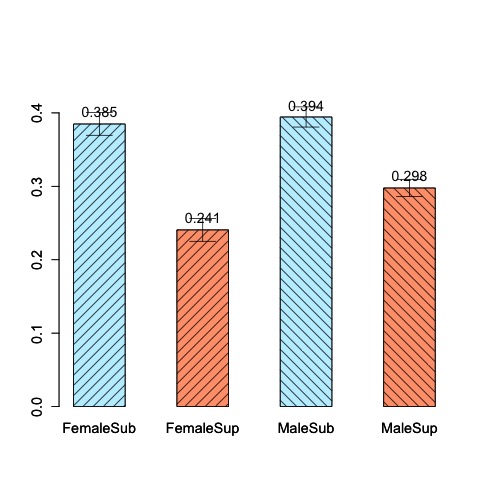}
	\caption{Mean value differences along Gender and Power: Initiator}
\small\small	(Error bars indicate standard error)
	\label{fig:FCAT_T_RDY_PST_AmItheInitiator}
\end{figure*}

\begin{figure*}
	\centering
	\includegraphics[width=0.4\linewidth]{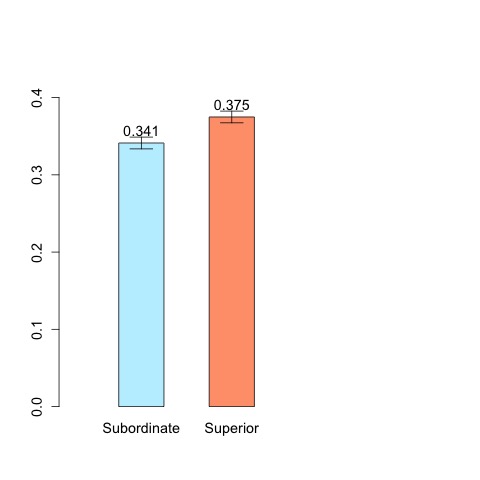}
	\hspace{-1.8cm}
	\includegraphics[width=0.4\linewidth]{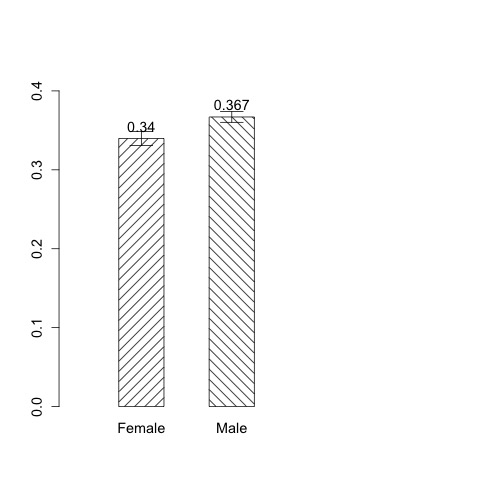}
	\hspace{-1.8cm}
	\includegraphics[width=0.4\linewidth]{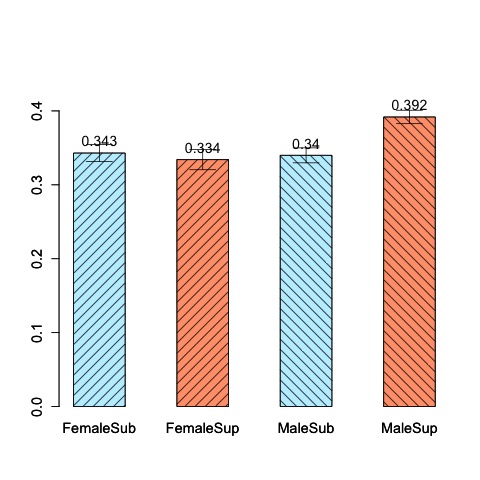}
	\caption{Mean value differences along Gender and Power:  LastMsgPos}
	\small	(Error bars indicate standard error)
	\label{fig:FCAT_T_RDY_PST_RelativePosOfLastMessage_HP}
\end{figure*}

Figure~\ref{fig:FCAT_T_RDY_PST_AmItheInitiator} 
shows the mean values of each groups for the feature \featInitiator.  
\featInitiator and {\featFirstMsgPos} behave more or less similarly; hence we show the chart only for \featInitiator.
Subordinates initiate the threads significantly more often than superiors (average value of 0.39 against 0.28 for \featInitiator). This pattern is also seen in \featFirstMsgPos (0.18 over 0.23; lower value means earlier participation). Both differences are highly statistically significant $ p<0.001 $. 
At first, this finding appears to be in contrast with our finding in \cite{prabhakaran-rambow:2014:P14-2} that superiors initiate more conversations.
As we discussed earlier, this is an artifact of the fact that broadcast messages with large number of recipients get eliminated from our corpus because it is more likely to fail to assign gender to at least one of the participants.
Putting together both findings, we infer that superiors tend to initiate email threads with large number of people; but in more focused conversations between smaller set of participants, it is the subordinates who initiate the conversations.

Gender is not a deciding factor. 
For \featInitiator, the t-Test result is significant ($ p=0.03 $), however the magnitude of difference is very small (0.32 for females over 0.34 for males; Figure~\ref{fig:FCAT_T_RDY_PST_AmItheInitiator}).
The t-Test result is not significant for \featFirstMsgPos. 
For the ANOVA test for the combination of gender and power, the result is not significant for \featInitiator.
The ANOVA test for \featFirstMsgPos is significant, however the Tukey's HSD test shows that 
male and female superiors behaved more or less the same way; similarly, male and female subordinates also behaved the same way.

The results on \featLastMsgPos is interesting (Figure~\ref{fig:FCAT_T_RDY_PST_RelativePosOfLastMessage_HP}). The t-Test results for both power and gender are significant, although the magnitude of the difference is small. The last message from superiors tend to come later than those of subordinates. Similarly, males tend to send their last messages later than females. The ANOVA results show that the factorial groups of power and gender also differ significantly ($ p<0.01 $). Upon Tukey's HSD test we find that male managers are the only group that differs from everyone else. The differences between all other groups are not statistically significant. But male managers differed from every other group significantly ($ p<0.01 $).
It is unclear why there is a significant difference in this feature. A potential explanation is that superiors tend to have the final word in conversations, and this is more in the case of male superiors. However, it is unclear to tease this apart as conversations are very often taken offline and hence it is hard to tell who had the final word. A more controlled study will need to be performed in order to verify this hypothesis, which we cannot perform using our corpus.

\subsection{Verbosity Features}
\label{sec:power_gender_vrb}

There are five features in this category --- \featMsgCount, \featMsgRatio, \featTokenCount, \featTokenRatio, and \featTokenPerMsg.
The first two features measure verbosity in terms of messages, whereas the third and fourth features measure verbosity in terms of words. The last feature measure how terse or verbose on average the messages are.

\begin{figure*}
	\centering
	\includegraphics[width=0.4\linewidth]{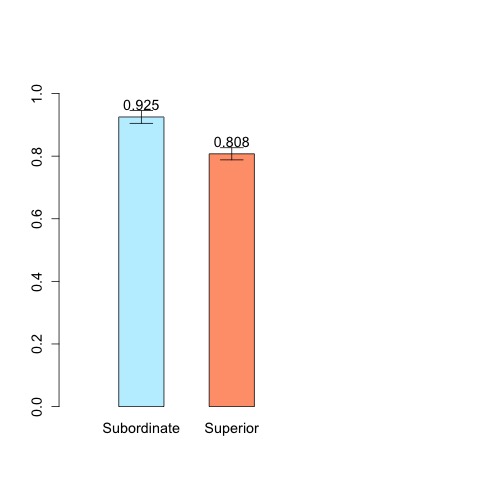}
	\hspace{-1.8cm}
	\includegraphics[width=0.4\linewidth]{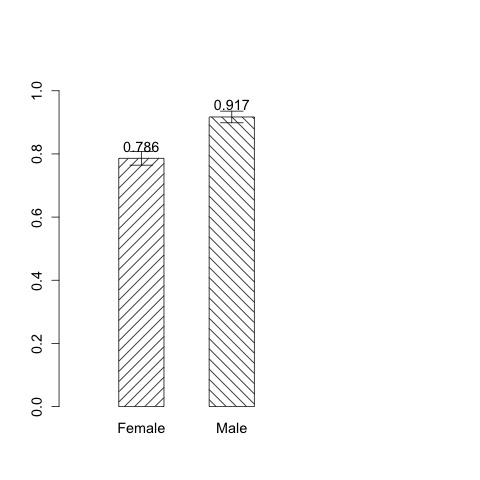}
	\hspace{-1.8cm}
	\includegraphics[width=0.4\linewidth]{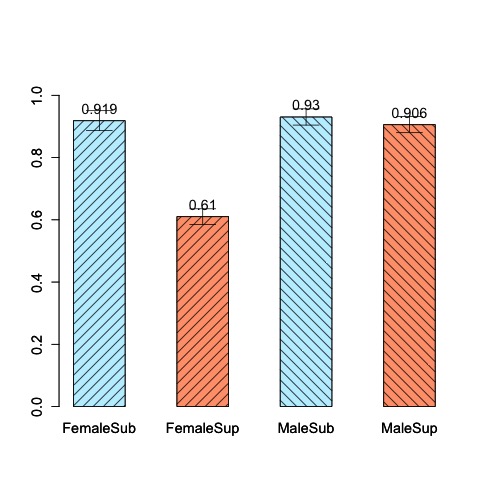}
	\caption{Mean value differences along Gender and Power:  MsgCount}
\small	(Error bars indicate standard error)
	\label{fig:FCAT_T_RDY_VRB_MessageCount}
\end{figure*}

\begin{figure*}
	\centering
	\includegraphics[width=0.4\linewidth]{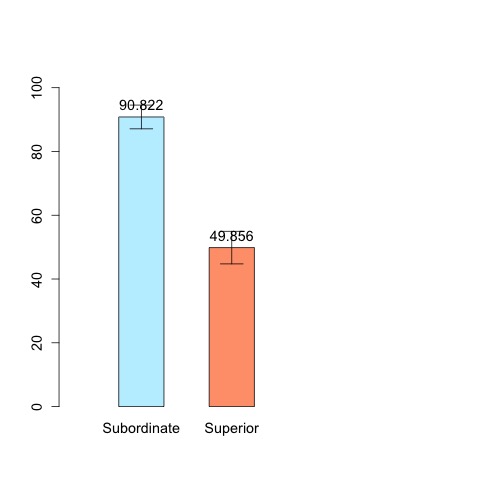}
	\hspace{-1.8cm}
	\includegraphics[width=0.4\linewidth]{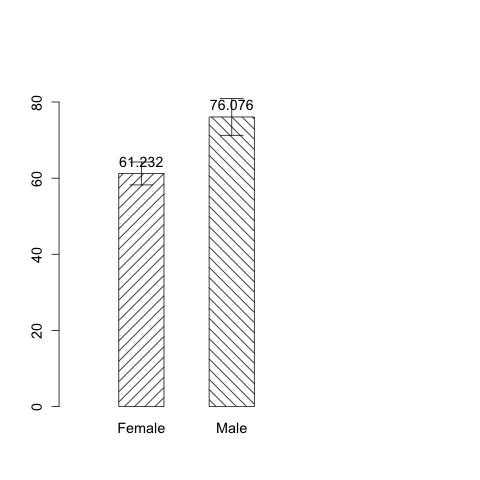}
	\hspace{-1.8cm}
	\includegraphics[width=0.4\linewidth]{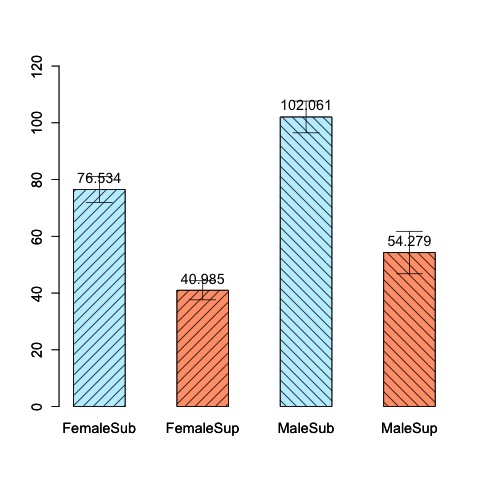}
	\caption{Mean value differences along Gender and Power:  TokenCount}
\small	(Error bars indicate standard error)
	\label{fig:FCAT_T_RDY_VRB_TokenCount}
\end{figure*}

\featMsgCount and \featMsgRatio behaved similarly, so did \featTokenCount and \featTokenRatio. 
Figure~\ref{fig:FCAT_T_RDY_VRB_MessageCount} and Figure~\ref{fig:FCAT_T_RDY_VRB_TokenCount} show the mean values of each groups for the feature \featMsgCount and \featTokenCount.
Superiors tend to send fewer of messages in the thread than subordinates ($ p<0.001 $), and women tend to send fewer messages than men ($ p<0.001 $). 
The ANOVA results for both \featMsgCount and \featMsgRatio are significant ($ p<0.001 $). Tukey's HSD test reveals an interesting picture. Female superiors send significantly fewer messages than everyone else, almost 25\% fewer than other groups.  In fact, they are the only single group that is different from anyone else. Difference between none of the other groups are significant. 
For \featTokenCount and \featTokenRatio, the results are similar.
Superiors tend to contribute fewer words in the thread than subordinates ($ p<0.001 $). Women tend to contribute fewer words than men ($ p<0.01 $). The ANOVA test of both features returned not significant.

\begin{figure*}
	\centering
	\includegraphics[width=0.4\linewidth]{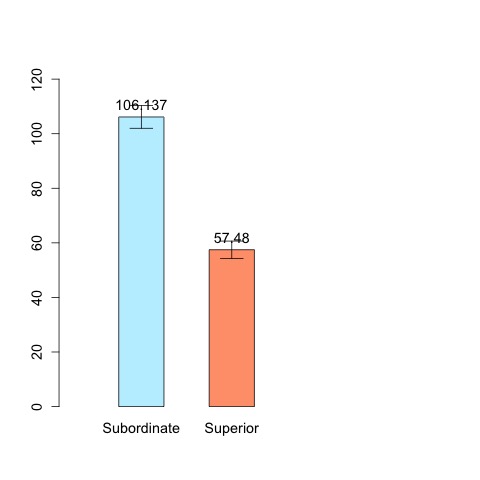}
	\hspace{-1.8cm}
	\includegraphics[width=0.4\linewidth]{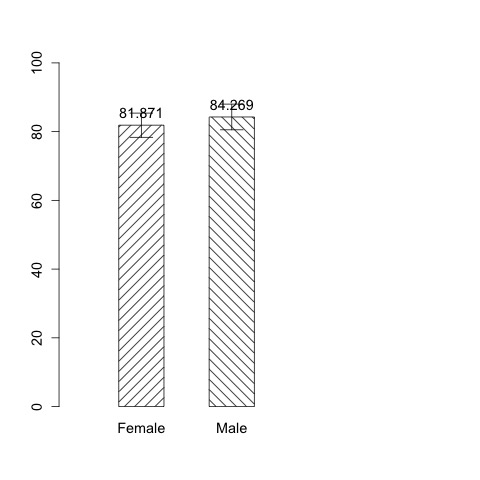}
	\hspace{-1.8cm}
	\includegraphics[width=0.4\linewidth]{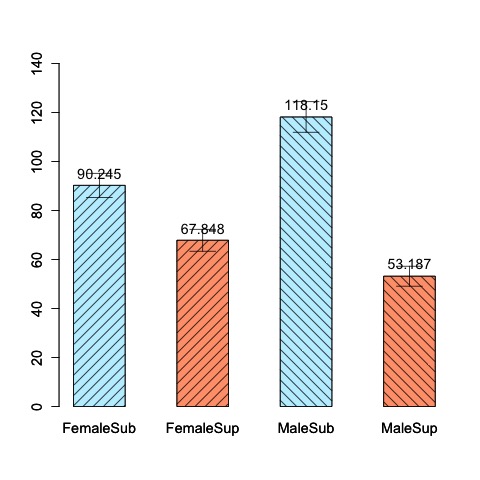}
	\caption{Mean value differences along Gender and Power:  TokenPerMsg}
\small	(Error bars indicate standard error)
	\label{fig:FCAT_T_RDY_VRB_TokensPerMessage}
\end{figure*}

\featTokenPerMsg behave differently. Gender is not significant at all. That is, men and women do not differ in how long their messages are. In terms of Power, subordinates send significantly longer emails. The ANOVA test is highly significant. It turns out that among superiors, there is no significant difference. But among subordinates, male subordinates send significantly longer emails than female subordinates ($ p<0.01 $) as per the Tukey's HSD test.  
In summary, power is a deciding factor in the difference between the verbosity exhibited by men and women. 
Female managers send significantly fewer messages than all other groups; both female and male managers send significantly shorter messages than subordinates. On the other hand, female subordinates send significantly shorter emails than male subordinates, although they do not differ in how many messages they send.

\subsection{Thread Structure Features}
\label{sec:power_gender_thr}

\begin{figure*}
	\centering
	\includegraphics[width=0.4\linewidth]{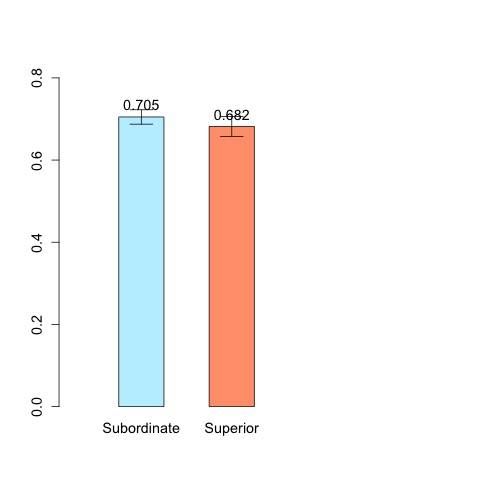}
	\hspace{-1.8cm}
	\includegraphics[width=0.4\linewidth]{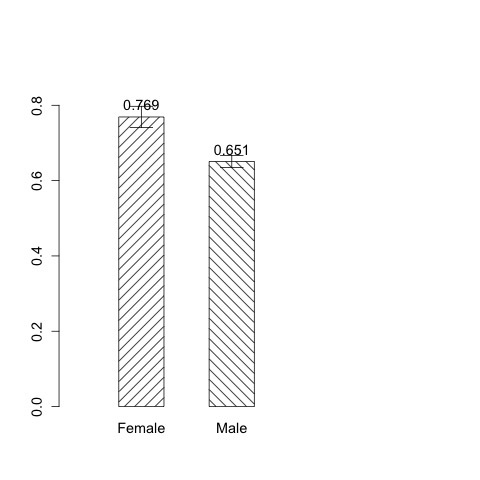}
	\hspace{-1.8cm}
	\includegraphics[width=0.4\linewidth]{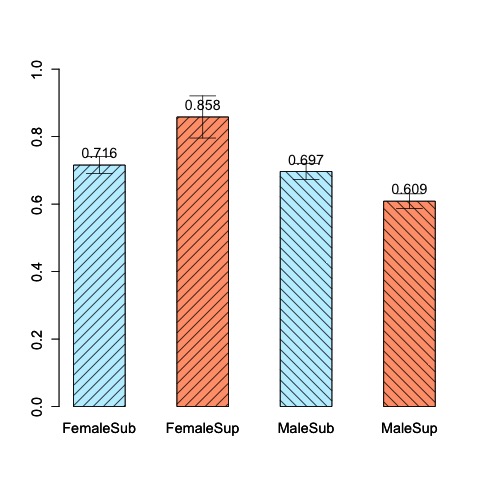}
	\caption{Mean value differences along Gender and Power:  ReplyRate}
	\small	(Error bars indicate standard error)
	\label{fig:FCAT_T_META_AverageReplyCount}
\end{figure*}

While the verbosity and positional features measure behavioral aspects, thread structure features in general deal with functional aspects (e.g., is a participant in CC (carbon copy) a lot?). While being in CC as a feature might be significantly related to power relations, it is unlikely that someone keeps a person in CC based on their gender. Similarly, adding or removing people to the conversation is also a functional aspect of workplace interactions, and we do not expect gender to play a role there. 
As expected there is no significant difference between women and men for \featInToListPercent, \featAddPerson, and \featRemovePerson.
The ANOVA test also returned not significant. In other words, gender does not affect the way superiors and subordinates behave in terms of these aspects.

\begin{figure*}
	\centering
	\includegraphics[width=0.4\linewidth]{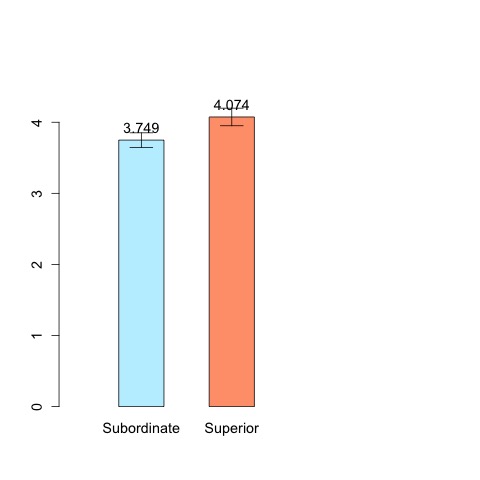}
	\hspace{-1.8cm}
	\includegraphics[width=0.4\linewidth]{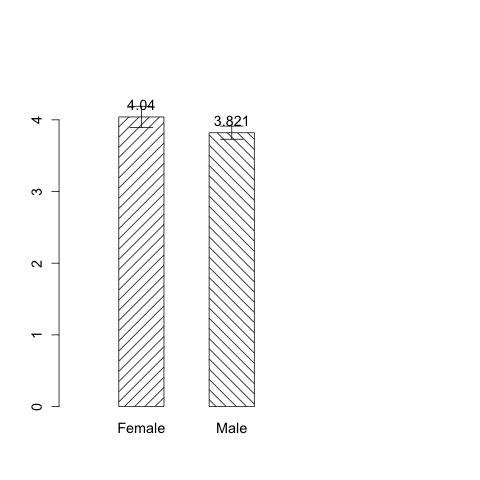}
	\hspace{-1.8cm}
	\includegraphics[width=0.4\linewidth]{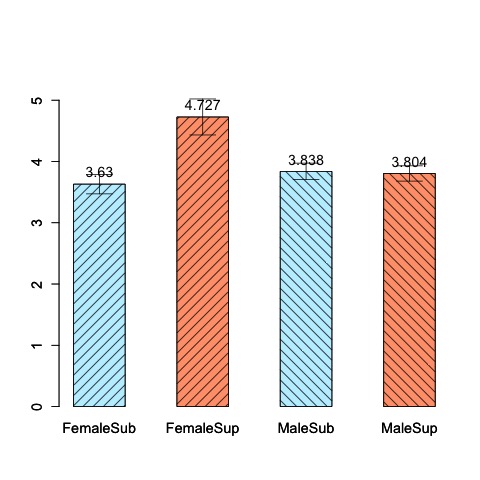}
	\caption{Mean value differences along Gender and Power:  AvgToRecipients}
	\small	(Error bars indicate standard error)
	\label{fig:FCAT_T_META_AvgeToPersons}
\end{figure*}

The results from our analysis of \featReplyRate is interesting. Figure~\ref{fig:FCAT_T_META_AverageReplyCount} shows the mean values for each group. Females get significantly more replies to their messages $ p<0.001 $. While power did not have a significant effect, the ANOVA result is also significant. On further analysis, we find that the female superiors get the highest reply rate ($ p<0.05 $). The difference between the \featReplyRate for male and female subordinates is not significant. 
It is an interesting finding, since it is an instance of gender of a person with power affecting how others behave towards them.
However, on combining this finding with the analysis of \featAvgRecipients and \featAvgToRecipients (Figure~\ref{fig:FCAT_T_META_AvgeToPersons}), we find that female superiors on average had more recipients in their messages than any other groups. The difference in \featReplyRate might also be a manifestation of the fact that female superiors send emails to larger number of people.

\begin{figure*}
	\centering
	\includegraphics[width=0.4\linewidth]{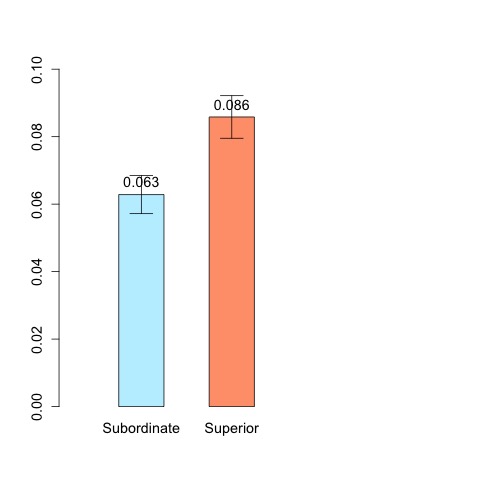}
	\hspace{-1.8cm}
	\includegraphics[width=0.4\linewidth]{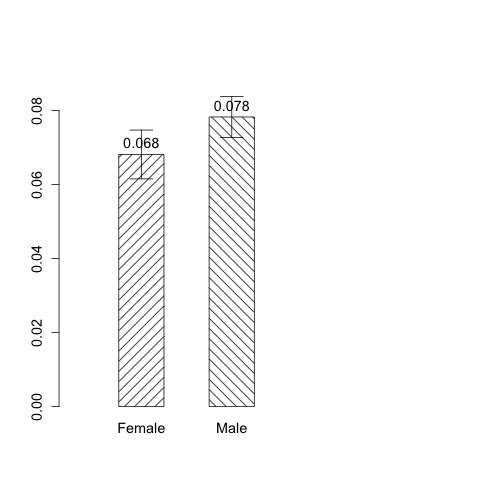}
	\hspace{-1.8cm}
	\includegraphics[width=0.4\linewidth]{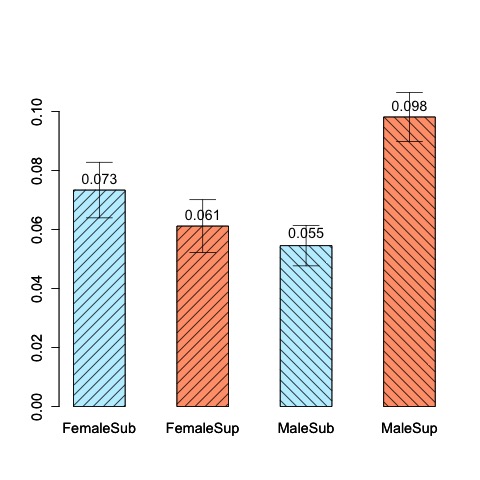}
	\caption{Mean value differences along Gender and Power:  ReqActionCount}
	\small	(Error bars indicate standard error)
	\label{fig:FCAT_T_TAG_DAP_DACount_Request_Action}
\end{figure*}

\subsection{Dialog Act Features}
\label{sec:power_gender_da}

We now discuss the finding in terms of dialog act counts.
\featInformCount and \featConventionalCount behave similarly for all three tests. However, the magnitude of difference between superiors and subordinates for \featInformCount is much higher than that of \featConventionalCount (superiors had 42.4\% lower value than subordinates for \featInformCount as opposed to 13.8\% in the case of \featConventionalCount).
The ANOVA test returned not significant, which means that the gender did not affect the way superiors or subordinates use either conventional or inform dialog acts.

On the other hand, the finding on 
\featReqActionCount and  \featReqInformCount are very interesting.
There is no significant difference between men and women in how often they make requests for action 
(Figure~\ref{fig:FCAT_T_TAG_DAP_DACount_Request_Action}), whereas they differed significantly ($ p<0.001 $) in terms of how often they request for information.
Women issue almost 41\% more requests for information than men.
The ANOVA test for \featReqActionCount returned significance ($ p<0.01 $), but not for \featReqInformCount. That is, gender affects how superiors and subordinates issue requests for actions, but not requests for information.
Male superiors issue more requests for actions than male subordinates, whereas female superiors held back from making requests. In fact, there is no significant difference between male subordinates and female subordinates in terms of \featReqActionCount.
For \featDanglingReqPercent, there is no significant difference with respect to gender or gender and power together.

\subsection{Overt Displays of Power}
\label{sec:power_gender_odp}

\begin{figure*}
	\centering
	\includegraphics[width=0.4\linewidth]{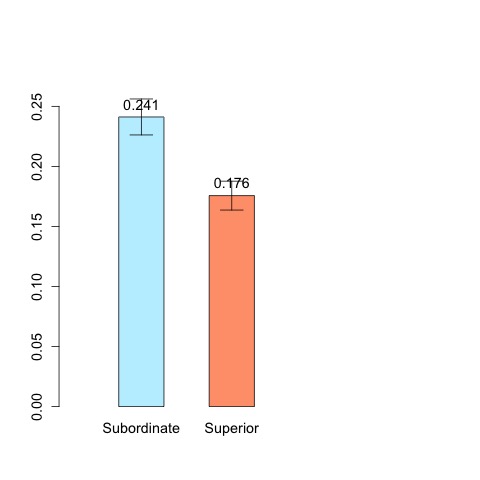}
	\hspace{-1.8cm}
	\includegraphics[width=0.4\linewidth]{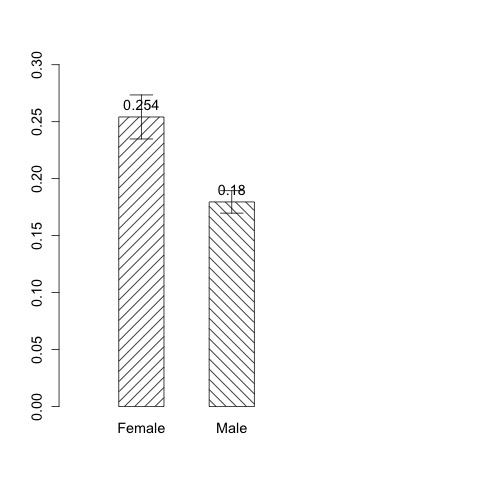}
	\hspace{-1.8cm}
	\includegraphics[width=0.4\linewidth]{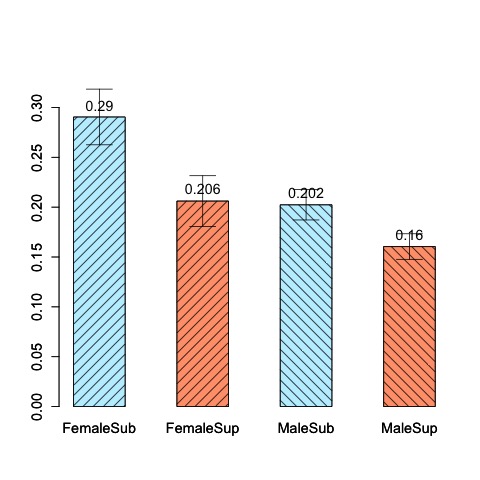}
	\caption{Mean value differences along Gender and Power:  ReqInformCount}
	\small	(Error bars indicate standard error)
	\label{fig:FCAT_T_TAG_DAP_DACount_Request_Information}
\end{figure*}

\begin{figure*}
	\centering
	\includegraphics[width=0.4\linewidth]{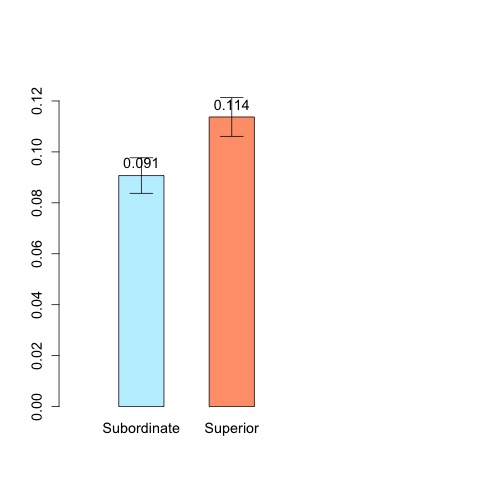}
	\hspace{-1.8cm}
	\includegraphics[width=0.4\linewidth]{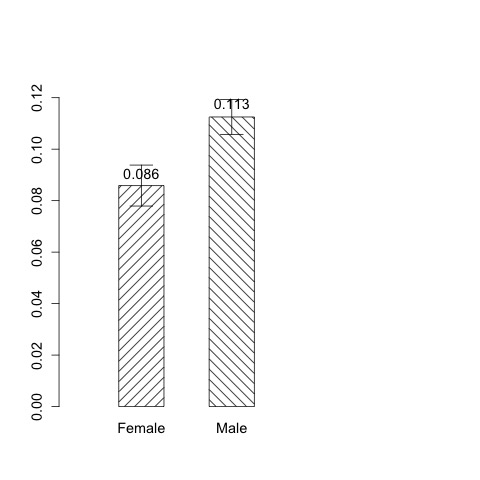}
	\hspace{-1.8cm}
	\includegraphics[width=0.4\linewidth]{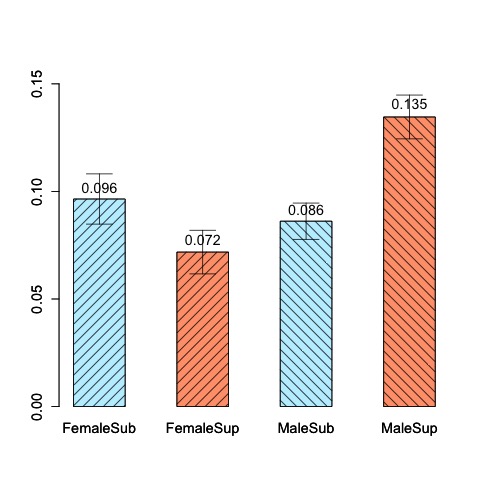}
	\caption{Mean value differences along Gender and Power:  ODPCount}
\small	(Error bars indicate standard error)
	\label{fig:FCAT_T_TAG_ODP_ODPCount}
\end{figure*}

Figure~\ref{fig:FCAT_T_TAG_ODP_ODPCount} shows the mean values of ODP counts in each group of participants.
The results obtained are similar to what we found for \featReqActionCount.
Both power and gender are significant on their own. 
Subordinates had an average of 0.091 ODP
counts and superiors had an average of 0.114 ODP counts. Gender is also
significant; females have an
average of 0.086 ODP counts and males had an average of 0.113 ODP counts. 
When looking at the factorial groups of power and gender, however, several differences are very highly
significant.  
Male superiors use the most ODPs, with an
average of 0.135 counts. Somewhat surprisingly, female superiors use the
{\em least} of the entire group, with an average of 0.072 counts.  
However, the differences among female superiors, female 
subordinates, and male subordinates are not significant, as per the Tukey's HSD test.

\subsection{Summary and Discussion}
\label{sec:summary_power_gender}

In summary, we find that gender affects the manifestations of power significantly along many linguistic and structural aspects of interactions. We summarize our findings below:

\begin{itemize}
	\item Gender of the participants does not have much effect on the manifestations of power in positional features (ref. Section~\ref{sec:power_gender_pst})
	\item Gender does significantly affect the manifestations of power in verbosity features; of the ANOVA tests we performed on the five verbosity features, three returned to be highly significant. (ref. Section~\ref{sec:power_gender_vrb})
	\item Gender also affects the manifestations of power on some of the thread structure features such as reply rate and number of recipients. (ref. Section~\ref{sec:power_gender_thr})
	\item Power manifestations on the dialog act based features, especially the request features and overt displays of power are also affected highly significantly by the gender of the participants. (ref. Section~\ref{sec:power_gender_da} and Section~\ref{sec:power_gender_odp})
\end{itemize}

\noindent The findings presented in this section do not exhaust the possibilities of this corpus. However, it shows how computational techniques can aid in performing large-scale sociolinguistics analysis. In order to demonstrate this point, we attempted to verify a hypothesis derived from the sociolinguistics literature we consulted. The hypothesis we investigate is:

\begin{itemize}
	\item
	
	{\bf Hypothesis 1}: Female superiors tend to use ``face-saving''
	strategies at work that include conventionally polite requests and
	impersonalized directives, and that avoid imperatives \cite{kendall200326}.
	
\end{itemize} 

\label{sec:face}

\noindent Our notion of overt display of power (ODP) is a face-threatening communicative strategy \cite{prabhakaranODP2012-long}.
An ODP 
limits the addressee's
range of possible responses, and thus threatens his or her (negative)
face.\footnote{For a discussion of the notion of ``face'', see
	\cite{brown_levinson_politeness}.}  We thus reformulate our hypothesis as
follows: the use of ODP by superiors changes when looking at the splits by
gender, with female superiors using fewer ODPs than male superiors.  
We saw in the results presented in Section~\ref{sec:power_gender_odp} that this hypothesis is indeed true.
We find that
female superiors used the
least number of ODPs among all groups.
The results confirmed our hypothesis: female superiors use fewer ODPs than
male superiors.  However, we also see that among women, there is no
significant difference between superiors and subordinates, and the
difference between superiors and subordinates in general (which is
significant) is entirely due to men.  This in fact shows that a more
specific (and more interesting) hypothesis than our original hypothesis is
validated: only male superiors use more ODPs than subordinates.
In other words, the fact that superiors use more ODPs than subordinates is entirely due to male superiors using more ODPs. Similarly, the fact that men use more ODPs than women is also entirely due to superiors among men using significantly more ODPs.

\section{Statistical Analysis: Gender Environment and Power}
\label{sec:gender_stat_genderenvpower}

In this section, we present our investigation on whether the manifestations of power differs based on the gender environment.
As in Section~\ref{sec:gender_stat_genderpower}, we use the ANOVA test to assess the statistical significance of differences. 
We perform ANOVA tests on all features 
keeping both Power and Gender Environment (GenderEnv, hereafter) as independent variables. We also perform ANOVA keeping GenderEnv alone as the independent variable; since GenderEnv has more than two groups, we cannot use Student's t-Test.
We verify our overall hypothesis that gender environment affects the way power is manifested in interactions; it still holds true even after applying the Bonferroni correction for multiple tests.
However, as we did in Section~\ref{sec:gender_stat_genderpower},  we do not apply the correction when describing the findings from the statistical analysis of each set of features separately in the rest of this section.

\subsection{Positional Features}
\label{sec:power_genderenv_pst}

\begin{figure*}
	\centering
	\includegraphics[width=0.4\linewidth]{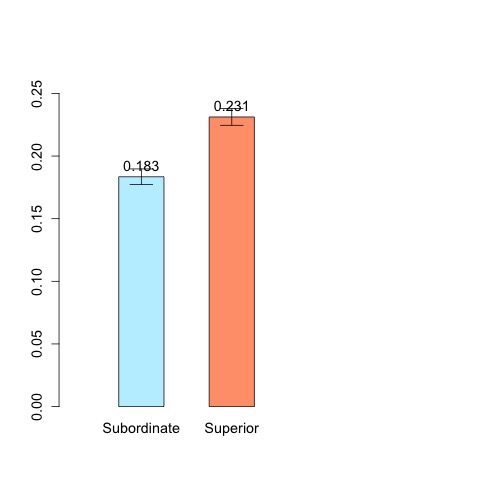}
	\hspace{-1.8cm}
	\includegraphics[width=0.4\linewidth]{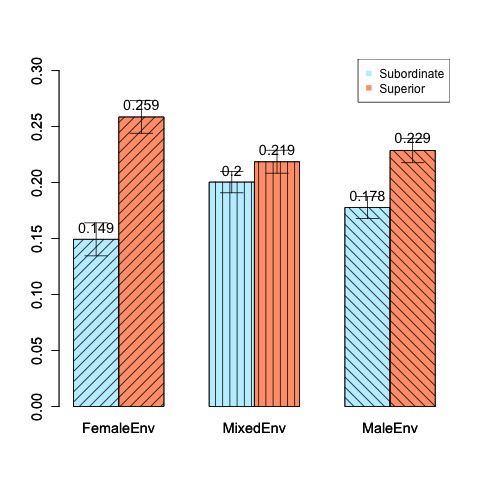}
	\caption{Mean value differences along Gender Environment and Power:  FirstMsgPos}
\small	(Error bars indicate standard error)
	\label{fig:FCAT_T_RDY_PST_RelativePosOfFirstMessage_Env}
\end{figure*}

For the positional features, any difference that we see in the feature values between different gender environments is not interesting. For example, it is not sensible to investigate whether the value of \featInitiator is different between gender environments (all threads had to be initiated by someone).
However, it is still interesting to see whether there is any connection between the gender environment and how the superiors and subordinates differ in terms of when they started and stopped participating in the threads.
As we saw in Section~\ref{sec:gender_stat_genderpower}, subordinates initiate more emails than superiors (\featInitiator) and overall start participating earlier in the thread (\featFirstMsgPos). The ANOVA test keeping Power and GenderEnv as independent variables was highly significant ($ p<0.001 $). In other words, the gender environment does affect the initiative shown by subordinates in starting email threads. Figure~\ref{fig:FCAT_T_RDY_PST_RelativePosOfFirstMessage_Env} shows the mean values of each group. 
Subordinates do start participating in the threads significantly earlier than superiors. However, the magnitude of this difference is dependent on the gender environment. 
This suggests that subordinates tend to show more initiative in female environments than other gender environments, and that superiors tend to start participating in the threads much later in female environments.
For the relative position of last message, the ANOVA results are not significant.

\subsection{Verbosity Features}
\label{sec:power_genderenv_vrb}

\begin{figure*}
	\centering
	\includegraphics[width=0.4\linewidth]{figures/3Group/FCAT_T_RDY_VRB_TokenCount/FCAT_T_RDY_VRB_TokenCount_HP}
	\hspace{-1.8cm}
	\includegraphics[width=0.4\linewidth]{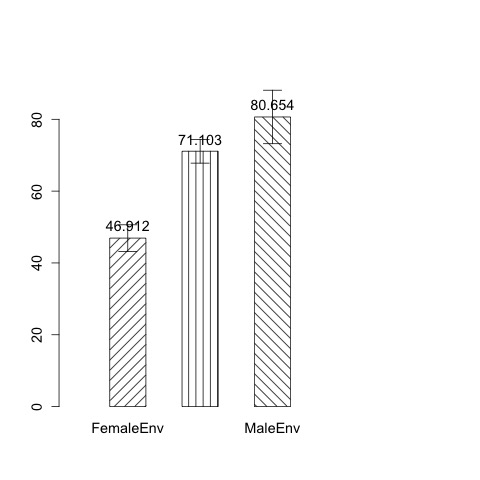}
	\hspace{-1.8cm}
	\includegraphics[width=0.4\linewidth]{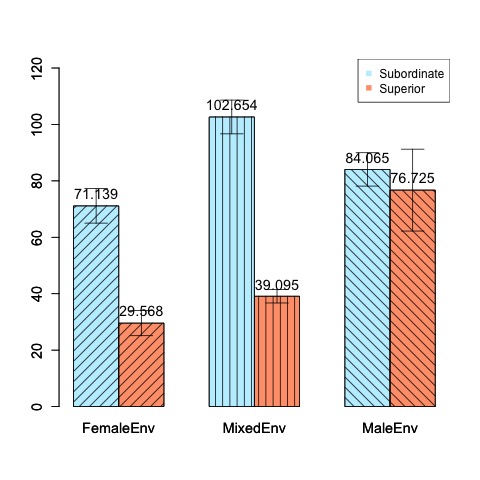}
	\caption{Mean value differences along Gender Environment and Power:  TokenCount}
\small	(Error bars indicate standard error)
	\label{fig:FCAT_T_RDY_VRB_TokenCount_Env}
\end{figure*}

As per the ANOVA results, the gender environment has no significance in \featMsgCount or in how Power is manifested in \featMsgCount. 
On the other hand, in terms of \featTokenCount, there is a significant difference ($ p<0.01 $) across gender environments (Figure~\ref{fig:FCAT_T_RDY_VRB_TokenCount_Env}).
The ANOVA test keeping Power and GenderEnv as independent variables also returned significance ($ p<0.001 $). In fact, in male environments, there is no significant difference in \featTokenCount between superiors and subordinates. 
Subordinates behaved more or less the same across the gender environments, but superiors contributed much less in female and mixed environments.
A similar pattern is also observed in \featTokenPerMsg across different gender environments.

\subsection{Thread Structure Features}
\label{sec:power_genderenv_thr}

The effect of gender environment on \featReplyRate is minimal. We observed that the number of recipients (both \featAvgRecipients and \featAvgToRecipients) is significantly higher in the mixed environment than others. This, however, is another artifact of how our corpus is constructed. In a thread with large number of participants, it is more likely to have a mixed environment than either male or female environment.
The ANOVA test keeping Power and GenderEnv also returned no significance for \featAddPerson and \featRemovePerson. In summary, the effect of gender environment on thread structure features is minimal.

\subsection{Dialog Act Features}
\label{sec:genderenv_power_da}

\begin{figure*}
	\centering
	\includegraphics[width=0.4\linewidth]{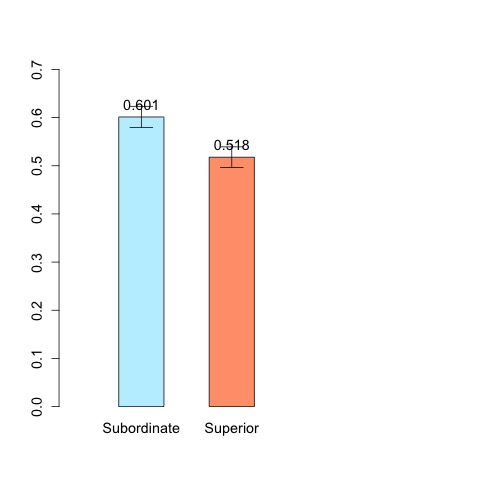}
	\hspace{-1.8cm}
	\includegraphics[width=0.4\linewidth]{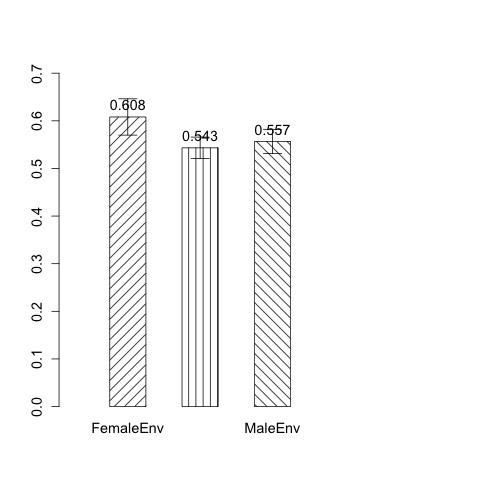}
	\hspace{-1.8cm}
	\includegraphics[width=0.4\linewidth]{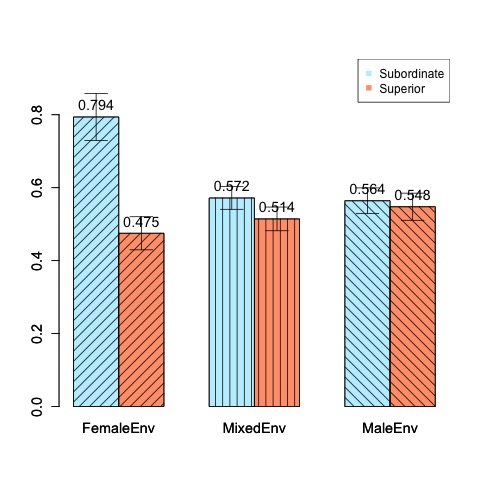}
	\caption{Mean value differences along Gender Environment and Power:  ConventionalCount}
\small	(Error bars indicate standard error)
	\label{fig:FCAT_T_TAG_DAP_DACount_Conventional_Env}
\end{figure*}

The results obtained on the ANOVA tests for the dialog act features are interesting. We will start with the \featConventionalCount.
Figure~\ref{fig:FCAT_T_TAG_DAP_DACount_Conventional_Env} shows the mean values of \featConventionalCount in each sub-group of participants.
Hierarchical Power is highly
significant as per ANOVA results. 
Subordinates use conventional language more (0.60) than superiors (0.52). 
While the averages by GenderEnv differ,
the differences are not significant. 
However, the groups defined by both
Power {\em and} GenderEnv have highly significant differences. 
Subordinates in female environments use the most
conventional language of all six groups, with an average of 0.79. Superiors
in female environments use the least, with an average of
0.48. 
In the Tukey HSD test, the only
significantly different pairs are exactly the set of subordinates in female
environments paired with each other group.
That is, subordinates in female environments use
significantly more conventional language than any other group, but the
remaining groups do not differ significantly from each other.
We interpret this result to mean that subordinates are more comfortable in female environments to use a style of communication which includes more conventional dialog acts than outside the female environments.

The ANOVA tests for \featInformCount also returned high significance. The difference between mean values of \featInformCount feature in male environments and mixed environments are not significant; but it differed significantly between female environments and both male and mixed environments.
The groups defined by both Power {\em and} GenderEnv also have highly significant differences. 
There is no significant difference between superiors' and subordinates' count of inform dialog acts when operating in a male environment. 
In other words, the finding that subordinates use more inform dialog acts holds true only in female and mixed environments, but not in male environments. However, on comparing this result with our findings in terms of verbosity features (Figure~\ref{fig:FCAT_T_RDY_VRB_TokenCount_Env}), we find that this is in fact an artifact of most of the contributions being inform statements (the findings in \featInformCount mirror that of \featTokenCount).

\begin{figure*}
	\centering
	\includegraphics[width=0.4\linewidth]{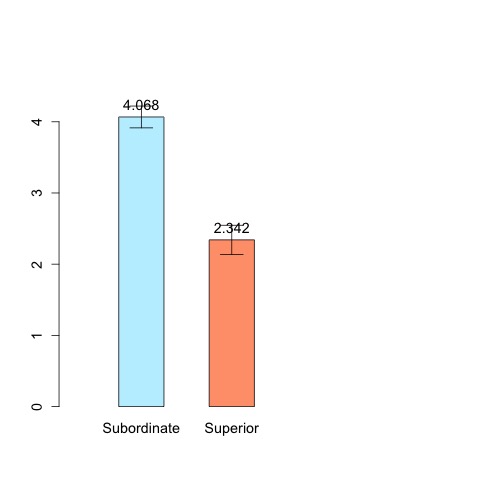}
	\hspace{-1.8cm}
	\includegraphics[width=0.4\linewidth]{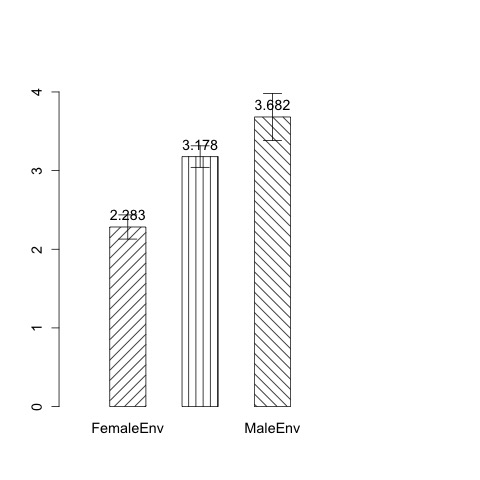}
	\hspace{-1.8cm}
	\includegraphics[width=0.4\linewidth]{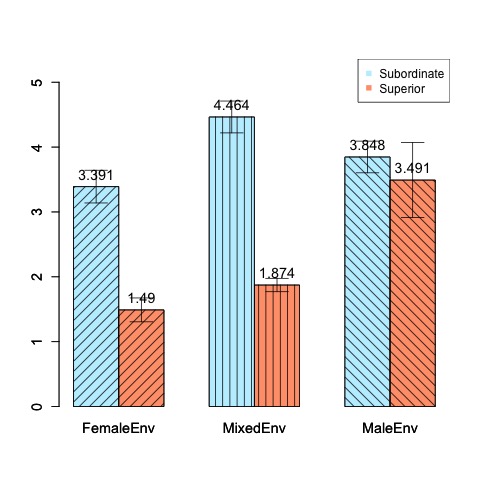}
	\caption{Mean value differences along Gender Environment and Power:  InformCount}
\small	(Error bars indicate standard error)
	\label{fig:FCAT_T_TAG_DAP_DACount_Inform_Env}
\end{figure*}

The ANOVA results for both \featReqActionCount, \featReqInformCount, and \featDanglingReqPercent are not significant when tested using Power \textit{and} GenderEnv. 
The male environment had a significantly ($ p<0.05 $) lower \featDanglingReqPercent.

\subsection{Overt Displays of Power}
\label{sec:power_genderenv_odp}

\begin{figure*}
	\centering
	\includegraphics[width=0.4\linewidth]{figures/3Group/FCAT_T_TAG_ODP_ODPCount/FCAT_T_TAG_ODP_ODPCount_HP}
	\hspace{-1.8cm}
	\includegraphics[width=0.4\linewidth]{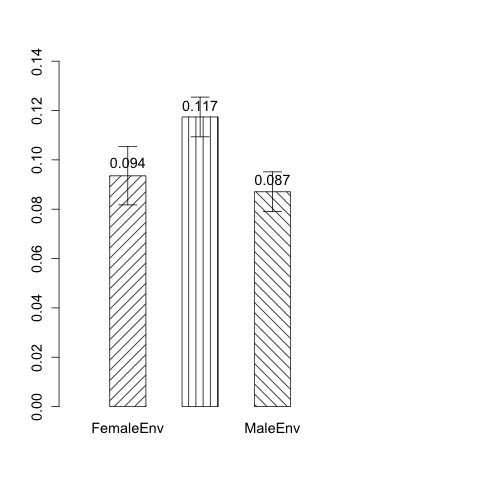}
	\hspace{-1.8cm}
	\includegraphics[width=0.4\linewidth]{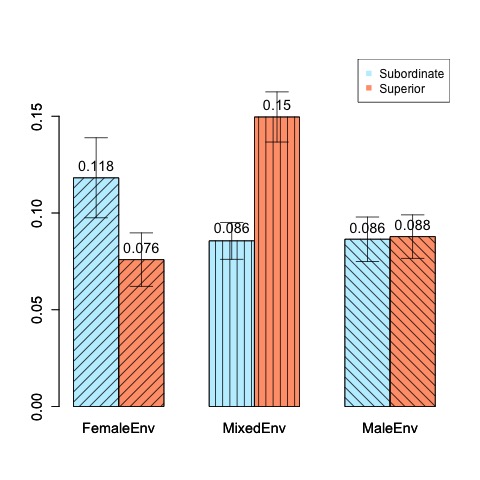}
	\caption{Mean value differences along Gender Environment and Power:  ODPCount}
\small	(Error bars indicate standard error)
	\label{fig:FCAT_T_TAG_ODP_ODPCount_Env}
\end{figure*}

The results of the ANOVA analysis on \featODPCount are interesting. 
Figure~\ref{fig:FCAT_T_TAG_ODP_ODPCount_Env} shows the mean values of each group.
As we saw already
in Section~\ref{sec:gender_stat_genderpower}, superiors use significantly more overt displays of power than subordinates. 
However, this pattern varied across gender environments significantly. 
The same relationship holds only in a mixed gender environment, where also most of the ODP occur.
In male environments, there is no significant difference in \featODPCount between superiors and subordinates, whereas in female environments, the value of \featODPCount for superiors is significantly lower than that of subordinates. This goes in line with our finding in Section~\ref{sec:gender_stat_genderpower} that female managers use fewer overt displays of power.

\subsection{Summary and Discussion}

In summary, we find that gender environment also affects the manifestations of power significantly along different structural aspects of interactions. We summarize the main findings below:
\begin{itemize}
	\item The gender environment significantly affects the difference between the initiative (in terms of how early they participated in the threads) exhibited by superiors and subordinates. While subordinates show more initiative than superiors across all gender environments, the magnitude of this difference is the largest in female environments. (ref. Section~\ref{sec:power_genderenv_pst})
	\item Gender environment affects the difference in verbosity exhibited by superiors and subordinates. While subordinates contributed significantly more content (in terms of token count as well as tokens per message) than superiors, this difference is the least in male environments. (ref. Section~\ref{sec:power_genderenv_vrb})
	\item Power manifestations on dialog act features also differ significantly across different gender environments. Subordinates use significantly more conventional dialog acts than superiors only in female environments. On the other hand, the difference in the their usage of inform dialog acts is non-existent in male environments. (ref. Section~\ref{sec:genderenv_power_da})
	\item Gender environment also affects the use of overt displays of power among subordinates and superiors. The fact that superiors use more overt displays of power is driven entirely by mixed environments. In male environments, superiors and subordinates do not differ in their usage of overt displays of power, while in female environments, superiors used less overt displays of power. (ref. Section~\ref{sec:power_genderenv_odp})
\end{itemize}

\noindent Similar to what we did in Section~\ref{sec:summary_power_gender}, we attempt to verify a hypothesis derived from the sociolinguistics literature we consulted in relation to the notion of gender environment. The hypothesis we investigate is:

\begin{itemize}
	
	\item
	
	{\bf Hypothesis 2}: Women when talking among themselves use language to create and maintain social relations, for example, they use more small talk (based on a reported ``stereotype'' in \cite{holmes2003feminine}). 
	
\end{itemize}  

We have at present no way of testing
for ``small talk'' as opposed to work-related talk, so we instead test Hypothesis 2 by asking how many conventional dialog acts a person performs.
Conventional dialog acts do not convey information or requests (both
of which would typically be work-related in the Enron corpus), but instead
establish communication (greetings) and to manage communication
(sign-offs); since communication is an important way of creating and
maintaining social relations, we can say that conventional dialog acts
serve the purpose of easing conversations and thus of maintaining social relations. 
We make our Hypothesis 2 more precise by saying that a higher
number of conventional dialog acts will be used in female environments.

We presented the results of our analysis of \featConventionalCount feature in Section~\ref{sec:genderenv_power_da}.
Our results first appears to be a negative result: while the averages by Gender Environment differ, the differences are not significant. 
However, we find that subordinates in female environments use
significantly more conventional language than any other group, but the
remaining groups do not differ significantly from each other.
Our hypothesis is thus only partially verified: 
while gender environment is
a crucial aspect of the use of conventional DAs, we also need to look at
the power status of the writer.  
While our hypothesis is not fully verified, we interpret the results to mean that subordinates are more comfortable in female environments to use a style of communication which includes more conventional DAs than outside the female environments.

\section{Utility of Gender Information in Predicting Power}
\label{sec:gender_powerpred}

\begin{table*}[t]
	\centering
	\captionsetup{justification=centering}
	\begin{tabular}{l l c  }
		\toprule
		& Description & Accuracy \\ 
		\midrule
		\multirow{1}{*}{Baselines}
		& Majority & 55.83 \\
		\midrule
		\multirow{2}{*}{Using gender features alone}
		& \genderShort & 57.59 \\
		& \genderShort + \genderEnvShort & 57.59 \\
		\midrule
		\multirow{4}{*}{Best feature sets}
		& \featSetNgramsShort + \featSetMetaDataShort + \genderShort + \genderEnvShort & \bf 70.74 \\
		& \featSetNgramsShort + \featSetMetaDataShort + \genderShort & 70.46 \\
		& \featSetNgramsShort + \featSetMetaDataShort & 68.24 \\
		& \featSetNgramsShort + \featSetMetaDataShort + \featSetPositionalShort + \featSetVerbosityShort & 68.33 \\
		\midrule
		\multirow{2}{*}{Best without \featSetNgrams}
		& \featSetDialogActsShort + \featSetOvertDisplayOfPowerShort + \featSetMetaDataShort + \genderShort & 67.31 \\
		& \featSetDialogActsShort + \featSetOvertDisplayOfPowerShort + \featSetMetaDataShort  & 64.63 \\
		\midrule
		\multirow{2}{*}{Best with no content}
		& \featSetPositionalShort + \featSetVerbosityShort + \featSetMetaDataShort + \genderShort  & 66.57 \\
		& \featSetPositionalShort + \featSetVerbosityShort + \featSetMetaDataShort  & 62.96 \\
		\bottomrule
	\end{tabular}
	\caption[Results on using gender features for power prediction]{\label{table:enron_gender_results}Results on using gender features for power prediction.}
	\featSetPositionalShort: \featSetPositional, 
	\featSetVerbosityShort: \featSetVerbosity, 
	\featSetMetaDataShort: \featSetMetaData, \\
	\featSetDialogActsShort: \featSetDialogActs, 
	\featSetOvertDisplayOfPowerShort: \featSetOvertDisplayOfPower,
	\featSetNgramsShort: \featSetNgrams, \\
	\genderShort: \gender
	\genderEnvShort: \genderEnv
\end{table*}

In this section, we investigate the utility of the gender information in the problem of predicting the direction of power presented in \cite{prabhakaran-rambow:2014:P14-2}.  We expect the SVM-based supervised learning system using quadratic kernel to capture the interdependence between dialog structure features and gender features that we found in our statistical analysis presented in Section~\ref{sec:gender_stat_genderpower} and Section~\ref{sec:gender_stat_genderenvpower}.

We perform our experiments on the \EnronGIEC subset, training a model using the same machine learning framework presented in  \cite{prabhakaran-rambow:2014:P14-2} using the related interacting participant pairs in the \textit{Train} subset of \EnronGIEC, and choosing the best model based on performance on the \textit{Dev} subset.
We experimented using all subsets of features described in Section~\ref{sec_features}. 
In addition, we add two gender-based feature sets: \gender containing the gender of both persons of the pair and \genderEnv which is a singleton set with the gender environment as the feature. 
Table~\ref{table:enron_gender_results} presents the results obtained using various
feature combinations.  
Note that the numbers presented in Table~\ref{table:enron_gender_results} are not directly comparable to the results presented in \cite{prabhakaran-rambow:2014:P14-2}, since the results presented there are on the \textit{Dev} set of the \EnronAll corpus, whereas here we discuss results obtained on the \textit{Dev} set of the \EnronGIEC, which is a subset of around 50\% of the \EnronAll corpus.

The majority baseline
obtains an accuracy of 55.8\%.
Using the gender-based features 
alone performs only slightly better than the majority baseline, posting an accuracy of 57.6\%.
The best performance is obtained using a combination of 
\featSetNgrams, \featSetMetaData, \gender and \genderEnv, which posts an accuracy of 70.7\%. Removing the \genderEnv feature set decreases the accuracy marginally to 70.5\%, whereas removing the \gender features as well reduces the performance significantly to 68.2\% (tested using McNemar test). 
This reduction of 2.4\% percentage points in accuracy shows that gender features are in fact useful for this power prediction task.  
The best performance feature set without using any gender information is the combination of 
\featSetNgrams, \featSetMetaData, \featSetPositional and \featSetVerbosity, which reports an accuracy of 68.3\%.
The best performing feature set without using \featSetNgrams is the combination of \featSetDialogActs, \featSetOvertDisplayOfPower, \featSetMetaData and \gender (67.3\%). Removing the gender features from this reduces the performance to 64.6\%.
Similarly, 
the best performing feature set which do not use the content of emails at all is \featSetPositional + \featSetVerbosity+ \featSetMetaData  + \gender (66.6\%). Removing the gender features decreases the accuracy by a larger margin (5.4\% accuracy reduction to 63.0\%).

It is interesting to look at the error reduction obtained by adding gender features to different feature sets. 
Using gender features alone obtains only an error reduction of 4.0\% over the majority baseline (i.e., without using any other features). 
However, the predictive value of gender features improves considerably when paired with other features.
For the best feature set we obtained, the gender features contributed to an error reduction of 7.9\% (68.2\% to 70.7\%). For the best feature set without using \featSetNgrams also the gender features contributed a similar error reduction of 7.6\% (64.63\% to 67.3\%). For the setting where no content features are used, gender features obtained an even higher error reduction of 11.0\% (63.0\% to 66.6\%).
In other words, the gender-based features
on their own are not very useful, and gain predictive value only when
paired with other features (as we are using a quadratic SVM kernel).
This is because the other features in fact
make quite different predictions depending on gender and/or gender
environment.  
Nonetheless, we take these results as validation of the claim that gender-based features enhance the
value of other features in the task of predicting power relations.

On our blind test set, the majority baseline obtains an accuracy of 57.9\% and
the baseline system that does not use gender features obtains an accuracy of 68.9\%.
On adding the gender-based features, the accuracy of the system improves to 70.3\%.

\section{Conclusion}
\label{sec:gender_conclusion}

The first contribution of this paper is the new, freely available resource --- Gender Identified Enron Corpus, an extension to the Enron email corpus with 87\% of the email senders' gender identified. We used the Social Security Administration's baby-names database to automatically assess the gender ambiguity of first names of email senders and assigned the gender to those whose names are highly unambiguous. Our gender identified corpus is orders of magnitude larger than other existing resources in this domain that capture gender information. 
We expect it to be a rich resource for
social scientists interested in the effect of power
and gender on language use.

Our second contribution is the detailed statistical analysis of the interplay of gender, gender environment and power in how they affect the dialog behavior of participants of an interaction. 
We introduced the notion of gender environment to capture the gender makeup of the discourse participants of a particular interaction. 
We showed that gender and gender environment affect the ways power is manifested in interactions
in complex ways, resulting in patterns in the discourse that reveal the underlying
factors.
While our findings pertain to the Enron email
corpus, we believe that the insights and techniques
from this study can be extended to other genres
in which there is an independent notion of hierarchical
power, such as moderated online forums.

Finally, we showed the utility of gender information in the task of predicting the direction of power between pairs of participants based on single threads of interactions. 
We obtained statistically significant improvements by adding the gender of both participants of a pair as well as the gender environment as features to a system trained using lexical and dialog structure features alone.

\section*{Acknowledgment}

This paper is partially based upon work supported by the DARPA DEFT program. The views expressed here are those of the author(s) and do not reflect the official policy or position of the Department of Defense or the U.S. Government. We thank Emily Reid who was involved in early stages of this work. We also thank Rob Voigt, Dan Jurafsky and the anonymous reviewers for their helpful feedback.

\bibliography{nl,vp}

\end{document}